\documentclass[pdflatex,sn-mathphys-num]{sn-jnl}

\usepackage{graphicx}%
\usepackage{multirow}%
\usepackage{amsmath,amssymb,amsfonts}%
\usepackage{amsthm}%
\usepackage{mathrsfs}%
\usepackage[title]{appendix}%
\usepackage{xcolor}%
\usepackage{textcomp}%
\usepackage{manyfoot}%
\usepackage{booktabs}%
\usepackage{algorithm}%
\usepackage{algorithmicx}%
\usepackage{algpseudocode}%
\usepackage{listings}%
\usepackage{pifont}
\usepackage{verbatim}
\usepackage{pifont}
\usepackage{multicol}
\usepackage{lscape}
\usepackage{float}      
\usepackage{placeins}   
\usepackage{capt-of}
\usepackage{graphicx}
\usepackage{hyperref}
\definecolor{gray1}{rgb}{0.95, 0.95, 0.96}

\newcommand{\xmark}{\ding{55}}%
\theoremstyle{thmstyleone}%
\theoremstyle{thmstyletwo}%

\theoremstyle{thmstylethree}%

\begin{document}

\title{NLP Datasets for Idiom and Figurative Language Tasks}


\author[1]{\fnm{Blake} \sur{Matheny}}\email{matheny.blake@jaist.ac.jp}

\author[1]{\fnm{Phuong Minh} \sur{Nguyen}}\email{phuongnm@jaist.ac.jp}

\author[1]{\fnm{Minh Le} \sur{Nguyen}}\email{nguyenml@jaist.ac.jp}

\author[2]{\fnm{Stephanie} \sur{Reynolds}}\email{s-reynolds@neptune.kanazawa-it.ac.jp}

\affil[1]{\orgdiv{Information Science}, \orgname{Japan Advanced Institute of Science and Technology}, \orgaddress{ \city{Ishikawa},  \country{Japan}}}

\affil[2]{\orgdiv{General Education}, \orgname{International College of Technology}, \orgaddress{ \city{Kanazawa},  \country{Japan}}}

\abstract{ 
Idiomatic and figurative language form a large portion of colloquial speech and writing. With social media, this informal language has become more easily observable to people and trainers of large language models (LLMs) alike. While the advantage of large corpora seems like the solution to all machine learning and Natural Language Processing (NLP) problems, idioms and figurative language continue to elude LLMs. Finetuning approaches are proving to be optimal, but better and larger datasets can help narrow this gap even further. The datasets presented in this paper provide one answer, while offering a diverse set of categories on which to build new models and develop new approaches. A selection of recent idiom and figurative language datasets were used to acquire a combined idiom list, which was used to retrieve context sequences from a large corpus.  One large-scale dataset of potential idiomatic and figurative language expressions and two additional human-annotated datasets of definite idiomatic and figurative language expressions were created to evaluate the baseline ability of pre-trained language models in handling figurative meaning through idiom recognition (detection) tasks. The resulting datasets were post-processed for model agnostic training compatibility, utilized in training, and evaluated on slot labeling and sequence tagging. 
}




\maketitle

\section{Introduction}\label{sec1}

The datasets created in this paper are downstream derivations of the Common Crawl \footnote{https://creativecommons.org/publicdomain/zero/1.0/} corpus. Many datasets have been derived from Common Crawl and shared for research. Current datasets focusing on idioms or figurative language have many features, but offer limited context, limited size, or outdated vernacular. In an effort to create a new, relevant, and updated dataset of idiomatic expressions and figurative language representing more recent vernacular, in this research OSCAR \footnote{https://oscar-project.org/} \cite{ortiz-suarez_sagot_romary_2019_asynchronous}and C4 \cite{C4raffel2023exploringlimitstransferlearning} filters of Common Crawl were used to match a compiled list of idioms. As an homage to the original researchers of OSCAR and C4, \textit{\textbf{p}otential} \textbf{i}diomatic and \textbf{f}igurative \textbf{l}anguage expressions (\textit{P}IFL), and IFL-OSCAR-A and IFL-C4-A, two additional human-annotated datasets of definite idiomatic and figurative language expressions were developed and used as training data for the task of idiomaticity detection. 

These idiomatic and figurative expressions are ubiquitous in natural language, but defy straightforward compositional interpretation. For language models, expressions like \textit{neither fish nor fowl} cannot be accurately understood by interpreting their individual words in isolation. As such, they pose significant challenges for NLP systems in tasks ranging from machine translation to semantic similarity and information retrieval.

Several idiomatic expression datasets have been introduced in recent years. LIdioms \cite{Moussallem2018}, EPIE\cite{saxena2020epie},  \cite{haagsma-etal-2020-magpie}, FLUTE \cite{chakrabarty-etal-2022-flute}, and SemEval-2022 \cite{tayyarmadabushi2022semeval2} provide annotated examples for various figurative language types, yet they differ in size, labeling schema, context availability, and idiom diversity. Consequently, a dataset was derived from OSCAR's filter of a snapshot of 2018 Common Crawl using phrase vocabulary combined from these datasets. The \textit{P}IFL-OSCAR dataset contains 5.8 million examples and 3,207 unique idioms. Social media has contributed to significant changes in the English lexicon. The increased in informality has begun permeating formal discourse. As idioms and figurative language appear more frequently during informal exchanges rather than formal situations, they appear more frequently in messages and thereby in lexicon used with AI chats. It is necessary to include as much current language as possible when evaluating and training for these sociolinguistic phenomena. In this work, our contributions are:

\begin{enumerate}
    \item  One large pre-sampled dataset for generalization.
    \item Two new annotated datasets with a larger variety of labels and metadata\footnote{The datasets and source code will be published upon acceptance.}.
    \item Cross evaluation on existing datasets.
    \item SOTA performance of sequence accuracy metric on MAGPIE.

    \end{enumerate}

Natural language understanding involves not only predicting words (token prediction) but interpreting roles (token class prediction), particularly when encountering figurative expressions. Idioms, which frequently defy compositional logic, require models to recognize spans that signal non-literal meaning and then assign them appropriate semantic labels or types. These idiomatic expressions are challenging due to their often unpredictable alignment with surface syntax and their cultural or contextual dependencies. Due to the abstract nature of many multi-word expressions which are typically non-compositional in nature, the combined semanticity is not equal to the parts/words in the phrase.  If the single meaning of a phrase is known, any small change in idiom phrasing will cause a language model to misclassify, misgenerate, or mistranslate (i.e., “trip up”). A native speaker will, according to Sinclair’s Idiom Principle \cite{sinclair1991ccc}, subconsciously excuse the incorrect wording and recognize the speaker’s or writer’s intended meaning. These small changes can be single word mistakes ("two birds with one rock/stone"),  or they can take the form of figures of speech errors, such as a mixed metaphor or malapropism, which can even be difficult for native speakers.  Common malapropisms include “one in the same” in the place of “one and the same”, “deep-seeded” for “deep seated”, and “could care less” for “couldn’t care less.” A mixed metaphor combines parts of two or more common expressions in an unusual way. An example of a mixed metaphor that has been observed in films is as follows, “people in glass houses sink ships.” The first part of this metaphor explains, “people in glass houses shouldn’t throw stones,” encouraging one not to criticize others for their faults. While the second part, “loose lips sink ships,” means to beware of unguarded talk. Humans tolerate and understand these mistakes, while a language model may not be robust to them.



\section{Related Work} 

This section represents a general overview of recent high-performing selection of models and methods, not a complete historical and comprehensive review of research on idiom and figurative language tasks and NLP applications.
Early improvements for phrase‐level representations in BERT \cite{devlin-etal-2019-bert} focused on contrastive learning and masked‐token tasks. Wang et al.\ introduced Phrase‐BERT, which fine‐tunes BERT with a paraphrase‐based contrastive objective to yield embeddings that better capture phrasal semantics and support applications like neural topic models through nearest‐neighbor phrase retrieval \cite{Wang2021PhraseBERT}. Qin et al.\ proposed IBERT , framing idiom detection as a cloze \cite{qin2021ibertidiomclozestylereading}. Similarly, Zhou et al.\ tackled idiomatic expression paraphrasing without strong supervision by combining unsupervised contextual signals with back-translation to build large‐scale parallel idiom–literal corpora, showing significant gains in BLEU, METEOR, and SARI for idiomatic sentence rewriting \cite{idiomaticexpression}.

A parallel approach leveraged contextualized word embeddings for disambiguation. Škvorc et al.\ introduced MICE, using ELMo \cite{elmo-peters-etal-2018-deep} and BERT embeddings as inputs to a neural classifier trained on a novel idiom dataset; they demonstrated cross‐lingual transfer and strong PIE detection even for unseen expressions \cite{_kvorc_2022}.  Kurfalı and Östling treated each candidate MWE as a single token and applied supervised and unsupervised classifiers over CWEs, achieving notable literal vs.\ idiomatic classification gains without extensive feature engineering \cite{kurfali-ostling-2020-disambiguation}. A STARSEM’22 study further showed that integrating static, masked‐token, and contextual embeddings from MLMs significantly boosts token‐level idiom classification in multilingual settings \cite{takahashi-etal-2022-leveraging}. Building on these insights, Yayavaram et al.\ introduced multi‐objective fine‐tuning of BERT with word cohesion and machine‐translation‐based contrastive losses, yielding state‐of‐the‐art sequence detection across seven diverse idiomaticity datasets \cite{yayavaram-etal-2024-bert}.

More recently, large language models (LLMs) have pushed re‐evaluation of idiomaticity detection. Phelps et al.\ benchmarked GPT-3, PaLM, and other LLMs on SemEval 2022, FLUTE, and MAGPIE, finding that while LLMs offer few-shot competitiveness, they consistently trail specialized, fine-tuned encoders \cite{phelps-etal-2024-sign}. De Luca Fornaciari et al.\ introduced IdioTS, a curated conversational LLM test suite for rare and novel idioms, revealing imperfect handling of unseen MWEs despite few-shot prompting strengths \cite{de-luca-fornaciari-etal-2024-hard}. Wu et al.\ analyzed idiomatic clusters in dialogue‐optimized LLM embeddings, showing more diffuse idiom representations compared to literal paraphrases and highlighting the need for embedding specialization \cite{donthi-etal-2025-improving}.

Specialized objectives have continued to advance well for figurative expressions: He et al.\ developed an adaptive contrastive triplet loss that mines hard positive/negative idiom samples, boosting SemEval 2022 Task 2 performance with asymmetric distance learning \cite{he-etal-2024-enhancing}. Zeng and Bhat modeled idiomaticity as semantic compatibility between expression and context, using an attention‐flow mechanism to fuse lexical and embedding features for robust PIE identification \cite{zeng-bhat-2021-idiomatic}. Madabushi et al.\ also released AStitchInLanguageModels, a bilingual MWE dataset with fine‐grained senses, demonstrating limited zero-shot but sample-efficient few-shot idiomaticity detection and representation learning  \cite{tayyar-madabushi-etal-2021-astitchinlanguagemodels-dataset}.

\section{Current Datasets}
High‐quality datasets are contributing to these advances: FLUTE frames idiomaticity as an NLI task with 1.7K idiom pairs explained via textual entailment \cite{chakrabarty-etal-2022-flute}; MAGPIE provides 56K crowd-annotated PIE instances from the BNC \cite{haagsma-etal-2020-magpie}; the EPIE static corpus supports cloze tasks and reading‐comprehension models \cite{qin2021ibertidiomclozestylereading}; LIDIOMS links multilingual idioms in RDF \cite{Moussallem2018};  and SemEval 2022 Task 2 delivers aligned detection and embedding splits in multiple languages \cite{tayyarmadabushi2022semeval2}. VNC \cite{VNC-cook-fazly-stevenson-2008-vnctokens} was previously a common benchmark for idioms, though after examination and consideration, it was omitted from this research. Surveying model behavior, Miletić and Schulte im Walde show that transformer encoders inconsistently capture MWE (multi-word expression) semantics—often relying on surface patterns and localizing semantics in early layers, revealing a lack of task specific pretraining ability\cite{miletic-walde-2024-semantics}. He et al.\ propose model-agnostic probing metrics (Affinity and Scaled Similarity) on noun‐compound minimal pairs, revealing persistent gaps in idiomaticity representation across static and contextual models \cite{he-etal-2024-enhancing}.  \textit{P}IFL-OSCAR and the subsequent samples have also been developed to offer new benchmarks for model development and evaluation. 

\begin{sidewaystable*}[htbp] 
\centering
\small
\caption{Summary of major idiom and figurative language datasets.}
\setlength{\tabcolsep}{3pt} 
\renewcommand{\arraystretch}{1.2} 
    \resizebox{\linewidth}{!}{%
    \begin{tabular}{|p{0.1\linewidth}|p{0.14\linewidth}|l|p{0.15\linewidth}|p{0.2\linewidth}|p{0.15\linewidth}|p{0.1\linewidth}|}
        \hline
        \textbf{Dataset} & \textbf{Discourse / Source Texts} & \textbf{Date Range} & \textbf{Annotation Type} & \textbf{Task / Training Setup} & \textbf{\# Idioms / Types} & \textbf{\# Entries / Instances} \\
        \hline
        \textbf{MAGPIE} &
        BNC: spoken, fiction, news, academic, non-fiction &
        1960s--1993 &
        Crowdsourced annotations for idiomatic vs literal vs other usage &
        Binary / multi-class idiomaticity detection; random vs type-based splits &
        $\sim$1,756 &
        $\sim$56,622 \\
        \hline
        \textbf{VNC-Tokens} &
        BNC (verb--noun combos in sentences) &
        1960s--1993 &
        Expert linguists, then validated &
        Binary classification (literal vs idiomatic) &
        53 &
        2,984 \\
        \hline
        \textbf{SemEval-2022 Task 2} &
        English (news, literature), Portuguese (OPUS), Galician (lit/news) &
        1990s--2010s &
        Crowdsourced, multilingual &
        Subtask A: idiomaticity classification; Subtask B: embeddings &
        English: 60; Portuguese: 50; Galician: 50 &
        $\sim$6,000 (all languages) \\
        \hline
        \textbf{EPIE} &
        BNC, COCA, newswire, conversational &
        1960s--2019 &
        Human annotation (literal vs idiomatic) &
        Sequence labeling: idiomatic vs literal spans &
        717 &
        $\sim$18,000--20,000 \\
        \hline
        \textbf{FLUTE} &
        Curated figurative texts: news, stories, online &
        2010s--2020s &
        Mix of crowdworkers + experts; explanation annotations &
        NLI framing (sentence pair entailment with figurative explanation) &
        4 categories (idioms, metaphors, similes, sarcasm) &
        $\sim$9,000 \\
        \hline
    \end{tabular}
    }
\label{tab:idiom_datasets}
\end{sidewaystable*}

The datasets utilized in this research represent outstanding work with unique annotation and compilation methods. In an effort to enhance idiom and figurative language research with the most updated type of language, Common Crawl filters were used. This not only diversifies the discourse type, but includes current vernacular that will more likely contain figurative language. EPIE and FLUTE show the most recent data. However, closer examination of the EPIE corpus revealed a large quantity of archaic syntax and lexicon. Since the direction of this research was recent contemporary lexicon, the EPIE corpus and dataset were omitted from training and testing, though its idiom vocabulary remained a principle component in building the new datasets. 

The existing state-of-the-art results for each relevant dataset are seen in Table \ref{tab:current_sota}. With the exception of the sentence-pair entailment task in FLUTE, MAGPIE is the most recently benchmarked dataset. MAGPIE is split into magpie{\_}random and magpie{\_}type{\_}aware.  The magpie{\_}random dataset was used in this research as a benchmark. FLUTE and SemEval 2022 Task 2 have been modified to be trained and evaluated with the slot tagging architecture in this paper. Only BIO labels were added in a new category, given the target phrase existence in the entry sequence. In Table \ref{tab:label_comparison}, the diverse categories of each dataset can be seen. With the exception of the specific figurative language type that exists only in FLUTE, the \textit{P}IFL-OSCAR and \textit{P}IFL-C4 datasets contain all diverse categories, intended for downstream model agnostic training and benchmarking. 

\begin{table*}[htbp] 
\caption{Current SOTA of idiom and figurative language datasets. The asterisk (*) denotes that the result is taken from the corresponding dataset paper.}
\resizebox{\linewidth}{!}{%

\begin{tabular}{l l l l }
\toprule
\textbf{Dataset} & \textbf{Task} & \textbf{Primary} & \textbf{Best} \\
\midrule
EPIE (Formal) & Span tagging (BIO) & Acc. & \textbf{98.00} *\\
MAGPIE (random) & Span id. (exact) & Sequence Accuracy& \textbf{91.51} \cite{yayavaram-etal-2024-bert}\\
MAGPIE (type-aware) & Span id. (unseen types) & Sequence Accuracy& \textbf{70.47} \cite{disc-zeng2021idiomatic}\\
\midrule
 SemEval-2022 2A & Idiomaticity existence& Macro-F1 (All)& \textbf{93.85} *\\
 FLUTE (Idioms)& Sentence-pair entailment& Accuracy& \textbf{79.20} *\\\bottomrule
\end{tabular}
}
\label{tab:current_sota}
\end{table*}

\begin{sidewaystable*}[htbp] 
\small
\centering
    \resizebox{\linewidth}{!}{%
\begin{tabular}{|p{2.3cm}|p{0.8cm}|p{1cm}|p{0.8cm}|p{1cm}
                |p{1cm}|p{1.cm}|p{1.2cm}|p{1.5cm}|p{1.5cm}|p{1.0cm}|p{1.3cm}|p{1.2cm}|}
\hline
\textbf{Dataset} & \textbf{prev/  next  sentence}& \textbf{para- phrase} &\textbf{BIO tags} & \textbf{POS tags} & \textbf{phrase type guess} & \textbf{length} & \textbf{position in sent} & \textbf{similarity metrics}  & \textbf{metadata} & \textbf{fig/lit label} & \textbf{Idiom Type Label} & \textbf{Human Annotated} \\
\hline
FLUTE & no & yes & no & no & no & yes & no & no & no & yes (all fig) & yes & yes \\
FLUTE + BIO (our addition) & no & yes & yes & no & no & yes & no & no & no & yes (all fig) & (as matched) & yes \\
SemEval 2022 & yes & no & no & no & no & no & no & no & no & yes (existence) & no & yes \\
SemEval 2022 + BIO (our addition) & yes & no & yes & no & yes & yes & yes & no & no & yes (existence) & (as matched) & yes \\
MAGPIE & yes & no & yes (0,1 only) & yes & yes & yes & yes & no & no & yes & no & (crowd-sourced) \\
OSCAR\textit{P}IFL (ours)& yes & no & yes & yes & yes & yes & yes & yes & yes & yes & no & no (direct match, possibles) \\
OSCAR\_IFL\_A (ours)& yes & no & yes & yes & yes & yes & yes & yes & yes & yes & no & yes \\
C4\_IFL\_A (ours)& yes & no & yes & yes & yes & yes & yes & yes & yes & yes & no & yes \\\hline 
\end{tabular}
}
\label{tab:label_comparison}
\end{sidewaystable*}

\section{Annotating Methodology}
The datasets introduced in this section contribute to the other primary resources for evaluating idiomatic and figurative language phenomena. Each dataset constructed was derived through a uniform semi-automated pipeline involving idiom retrieval, sentence extraction, and label augmentation with syntactic and semantic information. Post-processing steps incorporated POS tagging, BIO sequence labeling, cosine similarity metrics from BERT embeddings via the Span Search Algorithm, and contextual relationships across adjacent sentences. These enriched linguistic features enable diverse analytical and modeling approaches, including sequence tagging, contextual disambiguation, and representation learning. Following dataset construction, human annotation refined subsets of the data to ensure high-quality gold labels for figurative versus literal interpretation. The following subsections detail the composition and refinement of the \textit{P}IFL-OSCAR and \textit{P}IFL-C4 datasets into IFL-OSCAR-A and IFL-C4-A, while the results emphasize their scalability, linguistic diversity, and suitability for advancing research in idiom recognition and figurative language modeling.
\subsection{Pipeline Framework}
\begin{figure}[htbp] 
    \centering
    \includegraphics[width=1\linewidth]{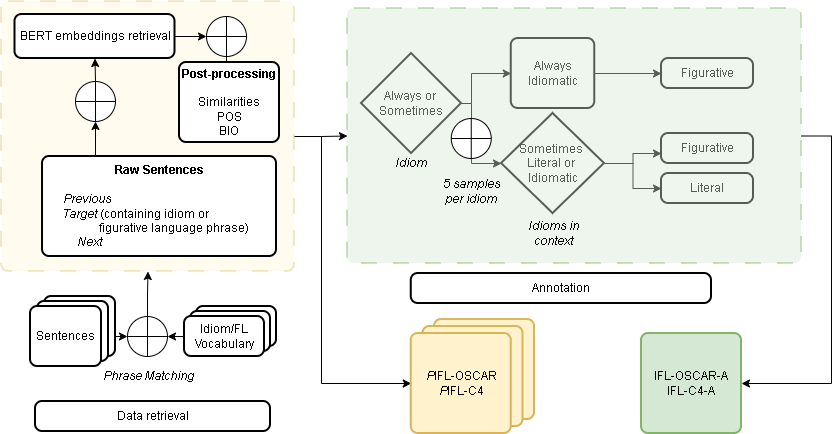}
    \caption{Data retrieval and annotation structure.}
    \label{fig:Pipeline}
\end{figure}

The raw dataset construction consisted of idiom and figurative language phrase vocabulary retrieval, Common Crawl filter choice, language choice,  phrase matching, sentence retrieval, and data compilation. Subsequent post-processing steps were performed to augment the raw data with POS tags, BIO labels, various statistics, and pre-computed similarities using BERT output embeddings. This completed the \textit{P}IFL-OSCAR dataset. 
After this semi-automated construction, human annotation was performed on a small sample. Beginning with a portion of the idiom list, it proceeded as follows and can be seen in Fig. \ref{fig:Pipeline}:
\begin{itemize}
    \item  Step 1: Collect idiom and figurative phrase vocabulary
    \item Step 2: Identify viable corpus
    \item Step 3: Retrieve raw target sentences and context via matching
    \item Step 4: Post process: cosine similarities via BERT output embeddings, POS tags, initial BIO labels
    \item Step 5: Classify a phrase as "Sometimes" or "Always" an idiom or figurative language
    \item Step 6: Choose "Literal" or "Figurative" for a sample of entries in the "Sometimes" set

\end{itemize}  

\subsection{Annotated Datasets}
\begin{figure}[htbp] 
    \centering
    \includegraphics[width=1\linewidth]{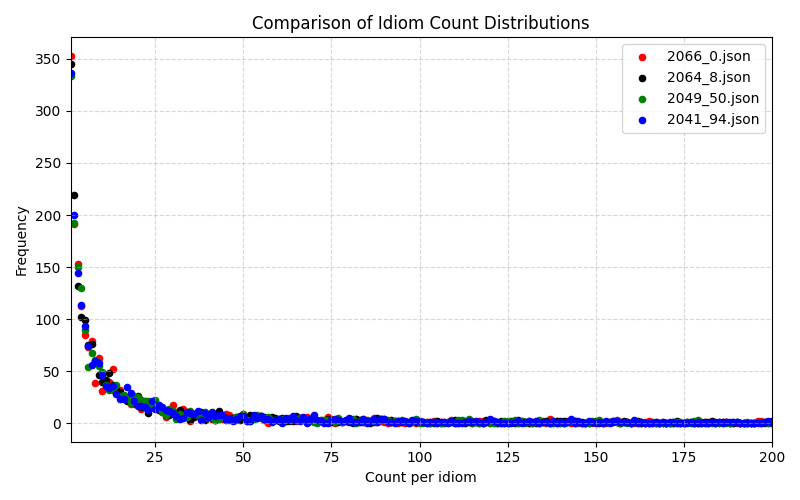}
    \caption{File distribution comparison showing a single file is a suitable sample of the collection.}
    \label{fig:sample_comparison}
\end{figure}

\paragraph{\textit{P}IFL-OSCAR}
The \textit{P}IFL-OSCAR dataset was built from a list of idioms derived from MAGPIE, FLUTE, EPIE, and LIdioms and matched with the OSCAR filter of the 2018 snapshot of Common Crawl. All were presumed to be possible idiomatic or figurative language. The number of idioms amounted to 5,789, though 3,209 matches occurred during initial extraction for the dataset. It initially consisted of 98 files of 7.7 million target sentences, amounting to ~21 GB. The data and metrics include POS tags for linguistic prior integration, BIO labels for sequence labeling, previous and next sentences for unsupervised learning, and many similarity metrics acquired from BERT output embeddings for post processing or supervised approaches. The dataset was further sanitized through keyword search to reduce specifically sexual, political, or discriminatory language, leaving a functional total of 5.83 million entries. Additionally, the list of unique idioms was reduced by the combination of similar phrases: "to be chuffed" = "be chuffed". 
By providing 97 subsets of the full dataset, it maintains the flexibility of training with samples. A sample of four subsets, named by the number of unique idioms and file number, were compared by idiom count distribution, shown in Fig. \ref{fig:sample_comparison}. This ensures that a single subset is an accurate sample of the full dataset of 97. 

\paragraph{\textit{P}IFL-C4}
A similar end-to-end extraction was performed, as illustrated in Fig. \ref{fig:Pipeline} for the C4-filter \cite{C4raffel2023exploringlimitstransferlearning} of the Common Crawl dataset\footnote{https://huggingface.co/datasets/allenai/c4}. This included phrase matching, BERT embedding processing, sanitizing via keyword, and combination of similar idioms. The subsequent annotation was also performed to acquire IFL-C4-A, though \textit{P}IFL-C4 was not used in testing due to its smaller size and its similar blanket figurativeness to \textit{P}IFL-OSCAR.

\subsection{Span Search Algorithm}

A re-imagining of an n-gram search algorithm was developed during preliminary testing of cosine similarity and figurative language testing. It created a set of phrases from a sentence. Such phrases were permutations of all sizes beginning at all positions. They were then ranked by the lowest cosine similarity of their output embeddings. For any Seq2Seq \cite{sutskever2014sequencesequencelearningneural} or masked language model that is designed to output token embeddings, the algorithm is:

For each sentence, $s$, of length $T$ with words, $w_t$, precompute its mean embedding
\[
 E_s = \frac{1}{T}\sum_{t=1}^T E_{w_t}.
 \]
We then slide a window of every length $\ell\in[2,L_{\max}]$ across positions $p=1,\dots,T-\ell+1$. Each span $(p,\ell)$ has a phrase embedding
\[
 E_p = \frac{1}{\ell}\sum_{i=0}^{\ell-1}E_{w_{p+i}},
\]
and a remainder embedding by subtracting the span’s contribution component-wise,
\[
 E_r = \frac{T\,E_s - \ell\,E_p}{T-\ell}.
\]
Compute two cosine similarities,
\[
 C_F = \cos(E_s, E_p), \quad C_R = \cos(E_r, E_p)\]
Preliminary testing showed a rank accuracy of exact phrase match for BERT-based models up to 10 percent as ranked by the lowest cosine similarity ranking of $C_R$  (phrase/sentence remainder). 
Augmenting the datasets with these types of pre-computed values allows future researchers to determine threshold values or design models based on these values. Previous and next sentence inclusion with their respective similarities to the target sentence align with SemEval idiom existence tasks and BERT next sentence prediction objectives. Conceptually, the cosine similarities that have the most influence on future target-sequence models are the phrase/full sentence and the phrase/sentence remainder values. These elucidate the contextual effect of the phrase on the sentence. In combination with the labels existing in current datasets, the post-processed data presents the ability to train a variety of models using:
\begin{enumerate}
    \item \textbf{BIO} - for sequence tagging, NER
\item \textbf{POS} - for parsing trees, linguistic priors
\item \textbf{Prev/Nex}t- for NSP (Next Sentence Prediction)
\item \textbf{Similarity} - for linguistic priors, composite scoring, candidate ranking
\item \textbf{Statistics} - n-gram modeling \footnote{A sample entry in Appendix 1.}

\end{enumerate}

\textbf{IFL-OSCAR-A and IFL-C4-A - Human Evaluation}
Most datasets established as benchmarks are heavily human annotated. For these datasets, two native-English-speaking linguists annotated figurative and literal labels for a subset of a single file of the \textit{P}IFL-OSCAR dataset. The IFL-C4-A dataset sample was annotated by this author only after confirming alignment with the other annotator for IFL-OSCAR-A. 
The human evaluation was implemented via a voting application developed in Gradio\footnote{https://www.gradio.app/}. Following the annotation pipeline, the initial votes on 300 idioms for the OSCAR-derived portion were voted "Always" and "Sometimes". 
Given two annotators choosing binary labels, Cohen's kappa coefficient was chosen to determine their alignment. Initial alignment was 82.6 percent.  "Always" votes were removed from the list. Next, 5 examples of each remaining "Sometimes" idiom were extracted from 3 separate files to ensure randomness. These were voted "figurative", "literal", "unsure", or "omit". The omit choice was included to account for specific entries that were not able to be filtered by the keyword sanitizing effort. Business names, offensive or discriminatory language, urls, and first-name personal references are examples of sentences that might be omitted. The "unsure" category and any disagreements were reviewed by both annotators and resolved. The dataset is a result of this process. After reviewing entries, the agreement was 100 percent. Table \ref{tab:OFL_stats} shows the final statistics of the datasets after omissions. IFL-OSCAR-A amounted to 695 entries with 152 unique idioms, while the IFL-C4-A contains 600 entries and 51 unique idioms. 

\begin{figure}[htbp] 
    \centering
    \includegraphics[width=1\linewidth]{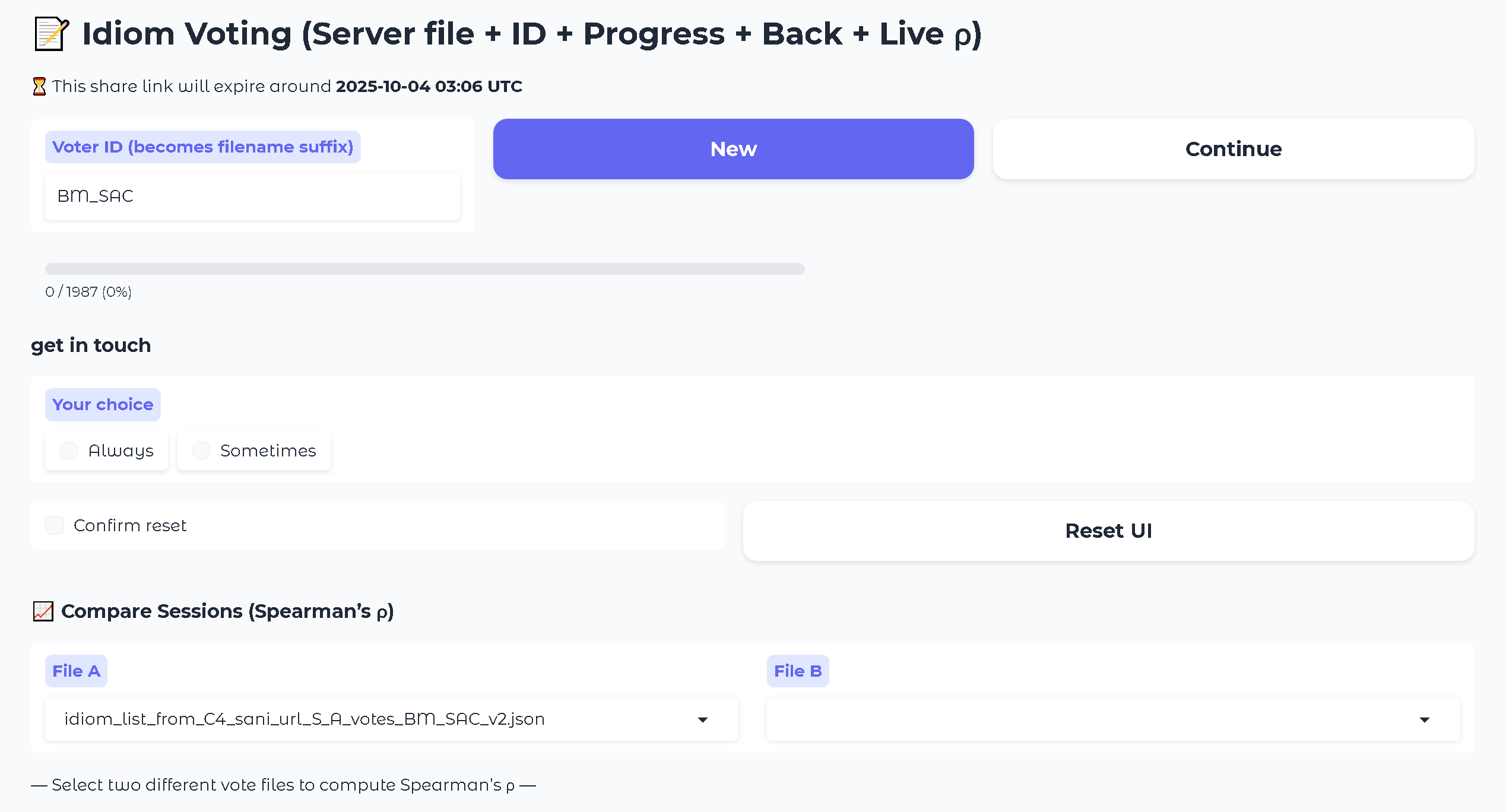}
    \caption{``Sometimes" and "always" initial annotation.}
    \label{fig:Sometimes_Always}
\end{figure}

\begin{figure}[htbp] 
    \centering
    \includegraphics[width=1\linewidth]{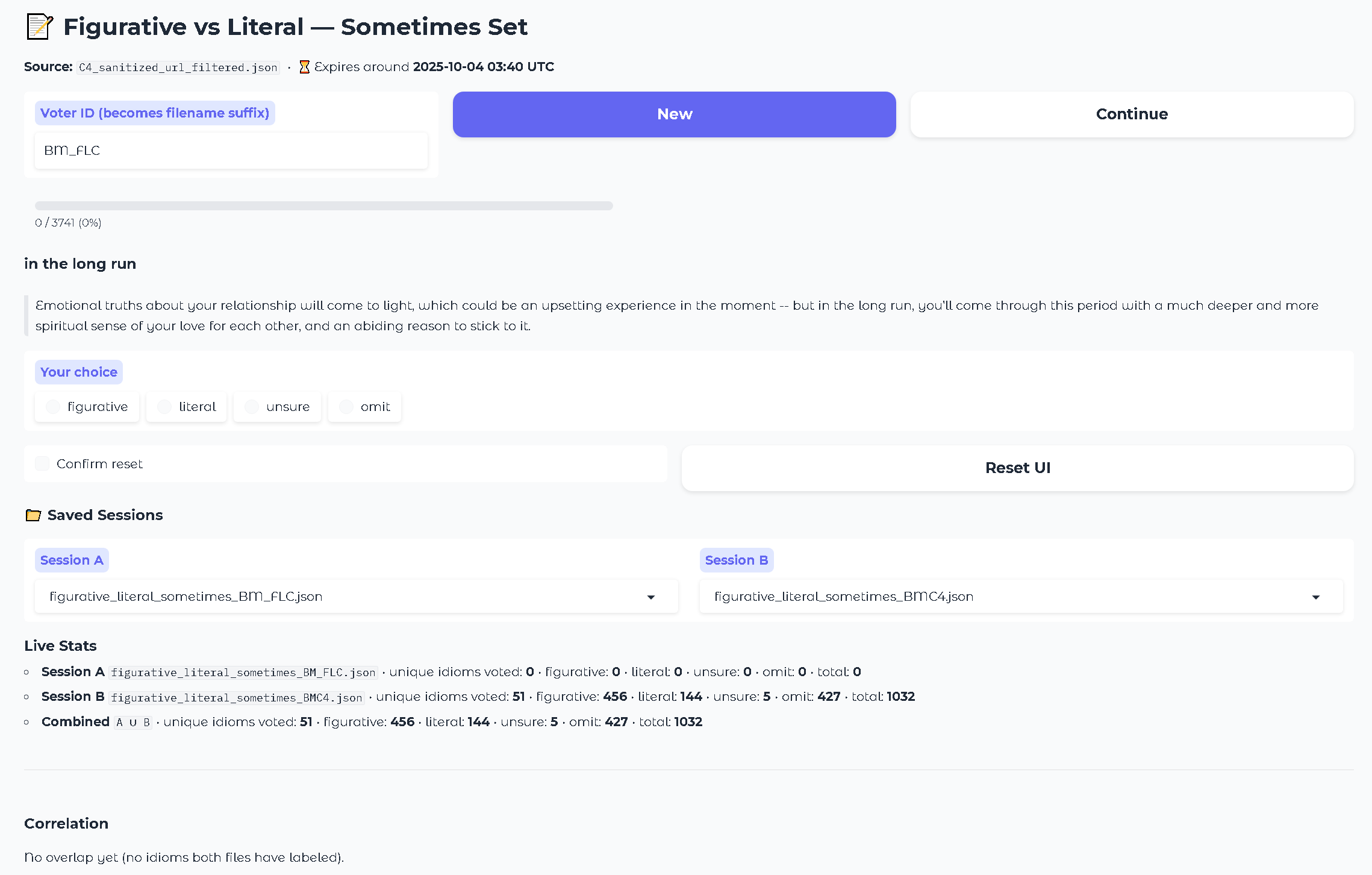}
    \caption{"Figurative" and "literal" annotation example.}
    \label{fig:Figurative_Literal}
\end{figure}

\begin{table*}[htbp] 
    \centering
    \caption{IFL-OSCAR-A statistics.} \label{tab:OFL_stats}
    \resizebox{0.99\linewidth}{!}{
    \begin{tabular}{|p{0.3\linewidth}|c|c|c|c|}\hline
         &  File 1 figurative&  File 1 literal&  File 2 figurative& File 2 literal\\
         Totals&  493&  210&  485& 249\\\hline
         Overlapping entries&  700&  agree: 646&  disagree: 54& Cohen's kappa: 82.36\\\hline
         After alignment \newline (and 3 omissions)&  & Figurative: 472&  Literal: 225& \\\hline
    \end{tabular}
    }
    
\end{table*}

\section{Experiments}

\subsection{Datasets}

  The datasets created in this pipeline contain a total of 1300 annotated examples and a substantial 5.8 million \textit{possible} entries, seen in Table \ref{tab:new_dataset_stats}. Not all examples in the larger datasets are figurative, though later discussion will show that it retains important utility.
\begin{table}[htbp]
\centering
\caption{Newly developed datasets and general statistics.}
\begin{tabular}{l r r r r}
\toprule
\textbf{Dataset} & \textbf{Entries} & \textbf{Unique Idioms} & \textbf{Figurative} & \textbf{Literal} \\
\midrule
\textit{P}IFL-OSCAR & 5,834,942 & 3152 & \textit{(possible IFL)} 5,834,942& 0 \\
\textit{P}IFL-OSCAR-HE & 700 & 152 & 471 & 226 \\
C4-PIFL & 3,741 & 728 & \textit{(possible IFL)} 3,741& 0 \\
 C4-PIFL-HE & 600 & 51 & 456 &144 \\
\bottomrule
\end{tabular}
\label{tab:new_dataset_stats}
\end{table}

The format of the newly created datasets contains a larger range of pertinent labels. The variety is intended for model-agnostic training and includes precomputed semantic relationships in addition to syntactical and statistical data. An example entry can be seen in Table \ref{tab_data_example}
\begin{table*}[htbp]
    \caption{One data entry from the \textit{P}IFL-OSCAR dataset. \label{tab_data_example}}
    \centering
    \resizebox{\linewidth}{!}{%
    \begin{tabular}{p{1.25\linewidth}}
    \noindent\colorbox{gray1}{\fbox{\parbox{1.\linewidth}{%
    \{
        \newline\hspace*{1em}\qquad\textbf{``labeling"}: \{
        \newline\hspace*{1em}\qquad\qquad\textbf{``seq.in"}: "Investing a little bit of time into research can save you a lot of money and trouble in the long run , and that is why you need to read this section carefully .",
        \newline\hspace*{1em}\qquad\qquad\textbf{``seq.out"}: "O O O O O O O O O O O O O O O O O B-idiom I-idiom I-idiom I-idiom O O O O O O O O O O O O O",
        \newline\hspace*{1em}\qquad\qquad\textbf{``BIO"}: ["O", "O", "O", "O", "O", "O", "O", "O", "O", "O", "O", "O", "O", "O", "O", "O", "O", "B", "I", "I", "I", "O", "O", "O", "O", "O", "O", "O", "O", "O", "O", "O", "O", "O"],
        \newline\hspace*{1em}\qquad\qquad\textbf{``seq.pos"}: "VBG DT JJ NN IN NN IN NN MD VB PRP DT NN IN NN CC NN IN DT JJ NN , CC DT VBZ WRB PRP VBP TO VB DT NN RB .",
        \newline\hspace*{1em}\qquad\qquad\textbf{``idiom\_POS"}: "IN"
        \newline\hspace*{1em}\qquad\},
        \newline\hspace*{1em}\qquad\textbf{``sentence\_similarities"}: \{
        \newline\hspace*{1em}\qquad\qquad\textbf{``target\_prev"}: 0.724046027407563,
        \newline\hspace*{1em}\qquad\qquad\textbf{``target\_next"}: 0.7110994006591008
        \newline\hspace*{1em}\qquad\},
        \newline\hspace*{1em}\qquad\textbf{``similarity\_metrics"}: \{
        \newline\hspace*{1em}\qquad\qquad\textbf{``pos\_pattern"}: "IN DT JJ NN",
        \newline\hspace*{1em}\qquad\qquad\textbf{``phrase\_type\_guess"}: "prepositional phrase",
        \newline\hspace*{1em}\qquad\qquad\textbf{``phrase\_tokens"}: ["in", "the", "long", "run"],
        \newline\hspace*{1em}\qquad\qquad\textbf{``sentence\_tokens"}: ["investing", "a", "little", "bit", "of", "time", "into", "research", "can", "save", "you", "a", "lot", "of", "money", "and", "trouble", "in", "the", "long", "run", "and", "that", "is", "why", "you", "need", "to", "read", "this", "section", "carefully"],
        \newline\hspace*{1em}\qquad\qquad\textbf{``idiom\_token\_span"}: [ 17, 18, 19, 20],
        \newline\hspace*{1em}\qquad\qquad\textbf{``cosine\_phrase\_sentence"}: 0.7851315550686879,
        \newline\hspace*{1em}\qquad\qquad\textbf{``cosine\_phrase\_minus\_phrase"}: 0.7110709750649673,
        \newline\hspace*{1em}\qquad\qquad\textbf{``cosine\_phrase\_left"}: 0.7259998505474241,
        \newline\hspace*{1em}\qquad\qquad\textbf{``cosine\_phrase\_right"}: 0.5783826655791272,
        \newline\hspace*{1em}\qquad\qquad\textbf{``cosine\_phrase\_local"}: 0.74453730902595,
        \newline\hspace*{1em}\qquad\qquad\textbf{``cosine\_delta\_main"}: 0.07406058000372062,
        \newline\hspace*{1em}\qquad\qquad\textbf{``contrast\_delta\_left"}: 0.05913170452126382,
        \newline\hspace*{1em}\qquad\qquad\textbf{``contrast\_delta\_right"}: 0.20674888948956072,
        \newline\hspace*{1em}\qquad\qquad\textbf{``idiomaticity\_score\_composite"}: 0.10350043850456644,
        \newline\hspace*{1em}\qquad\qquad\textbf{``phrase\_sentence\_cosine"}: 0.5259013175964355,
        \newline\hspace*{1em}\qquad\qquad\textbf{``backend"}: "hf-bert",
        \newline\hspace*{1em}\qquad\qquad\textbf{``model\_name"}: "bert-base-uncased"
        \newline\hspace*{1em}\qquad\},
        \newline\hspace*{1em}\qquad\textbf{``statistics"}: \{
        \newline\hspace*{1em}\qquad\qquad\textbf{``sentence\_token\_count"}: 32,
        \newline\hspace*{1em}\qquad\qquad\textbf{``idiom\_token\_count"}: 4,
        \newline\hspace*{1em}\qquad\qquad\textbf{``idiom\_position\_start"}: 17,
        \newline\hspace*{1em}\qquad\qquad\textbf{``idiom\_position\_ratio"}: 0.53125,
        \newline\hspace*{1em}\qquad\qquad\textbf{``ratio\_idiom\_to\_sentence\_length"}: 0.125,
        \newline\hspace*{1em}\qquad\qquad\textbf{``percentage\_idiom\_position\_in\_sentence"}: 53.125
        \newline\hspace*{1em}\qquad\},
        \newline\hspace*{1em}\qquad\textbf{``line\_index"}: 15,
        \newline\hspace*{1em}\qquad\textbf{``model"}: \{
        \newline\hspace*{1em}\qquad\qquad\textbf{``name"}: "bert-base-uncased",
        \newline\hspace*{1em}\qquad\qquad\textbf{``embedding\_layer\_type"}: "last\_4\_mean"
        \newline\hspace*{1em}\qquad\},
        \newline\hspace*{1em}\qquad\textbf{``match\_type"}: "exact string match",
        \newline\hspace*{1em}\qquad\textbf{``phrase"}: "in the long run",
        \newline\hspace*{1em}\qquad\textbf{``phrase\_canonical"}: "in the long run",
        \newline\hspace*{1em}\qquad\textbf{``phrase\_variants"}: [
        \newline\hspace*{1em}\qquad\qquad"in the long run"
        \newline\hspace*{1em}\qquad],
        \newline\hspace*{1em}\qquad\textbf{``phrase\_representative"}: "in the long run"
        \newline\hspace*{1em}\}
        }}}
         \\
    \end{tabular}
    }
\end{table*}

\subsection{Experimental Models}

\begin{figure}[htbp] 
    \centering
    \includegraphics[width=0.55\linewidth]{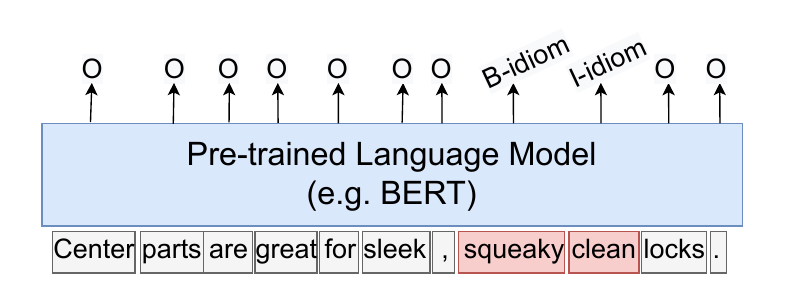}
    \caption{BERT-based model utilizing sequence classification for idiom recognition tasks. The red highlight indicates the idiom phrase in the sentence.}
    \label{fig:bertbased_architecture_id_task}
\end{figure}

Baseline testing of these datasets was conducted with a sequence labeling architecture. As a labeling architecture, it relied on parallel files of words and labels. An additional file containing a set of labels was required for word label prediction. The architecture, whose code base was re-implemented from JointBERT CAE \cite{phuong2022cae} via Chen, et al \cite{chen2019bert-intent-slot} for joint loss modeling,  was set up to accept various types of BERT-based models and a selection of other LLMs. 
\paragraph{BERT-based Method.} The BERT-based model architecture comprises multiple Transformer encoder layers. Each encoder layer contains a self-attention mechanism that serves as the primary component for extracting and modeling linguistic features. This mechanism enables the model to capture long-range dependencies between word pairs within a sentence. Given an input sentence ($\mathbf{s} = \{w_i\}_1^{|s|}$ where $|s|$ denotes the number of words\footnote{in practice, the input is tokenized into subwords using the BERT tokenizer.}, the BERT model encodes each token into its corresponding hidden vector representation ($\mathbf{h}_i$). 
    \begin{align}
        \mathbf{h}^{[CLS]}, \mathbf{h}^{word}_i &= \mathrm{BERT}( \mathbf{s} ) 
    \end{align}
Following \citet{devlin-etal-2019-bert},  the idiom recognition task is approached using a sequence labeling architecture (see Fig.~\ref{fig:bertbased_architecture_id_task}). After obtaining the contextualized word representations $\mathbf{h}^{word}_i$ from the BERT encoder, each vector is passed through a fully connected (Dense) layer and a \textit{softmax} classifier:
 
            \begin{align}
                \hat{\mathbf{y}}_i = \mathrm{softmax}( \mathbf{W}\mathbf{h}_i^{word} + \mathbf{b}  ) \label{eq:sf}
            \end{align}
            where  $\mathbf{W}, \mathbf{b}$ is learnable parameters. When a word is segmented into multiple subwords by the BERT tokenizer, only the first subword representation is used to represent the entire original word.
For model training, the objective function is defined as the weighted sum of cross-entropy losses across all tokens in the sequence:
            losses of SF and ID sub-tasks. 
            \begin{align}
                \mathcal{L} = \frac{1}{|s|} \sum_{i=1}^{|s|} \mathrm{CrossEntropy}(\hat{\mathbf{y}}_i, \mathbf{y}_i ) 
            \end{align}
            where $\mathbf{y}_{i}$ is the gold labels from Idiom Recognition datasets. 

\paragraph{LLM-based Method.} The ability of LLMs in the idiom recognition task was evaluated using a zero-shot instruction prompting technique. The selected LLMs were based on a Transformer decoder–only architecture. These models include Gemma-3, Llama-3.1, GPT-OSS, Mistral, Phi-4, and Qwen-2.5. Baseline testing was also performed on large chat-based LLMs via Ollama \footnote{
Open-weight models (Gemma-3, Llama-3.1, GPT-OSS, Mistral, Phi-4, and Qwen-2.5) were accessed locally through the \textit{Ollama} framework 
(\url{https://ollama.com}). See model-specific citations \citep{meta2024llama3_1,mistral2023mistral7b,deepmind2025gemma3,microsoft2024phi4,qwen2025qwen2_5}.
}.
Given a sentence, the prompting template shown in Table~\ref{tab_instruction_prompting} was used to instruct the LLM to understand and identify the idiom phrase in the sentence.
\begin{align}
\mathrm{output} = \mathrm{LLM}(\mathrm{prompting}(\mathbf{s}))
\end{align}
where \textit{LLM} denotes the corresponding language model, and the \textit{prompting} procedure applies the instruction template to explain the idiom recognition task to the model for a given input sentence~$\mathbf{s}$.

\begin{table}[htbp]
    \caption{Instruction prompting template for LLM-based idiom recognition tasks. }\label{tab_instruction_prompting}
    \centering
    \begin{tabular}{p{0.94\linewidth}}
    \noindent\colorbox{gray1}{\fbox{\parbox{1.\linewidth}{%
        You are a slot tagger. Use ONLY the following tag types: \newline
        Use ONLY prefixes `B' and `I'. \newline
        If a token is outside, use `O'. \newline
        Return exactly N tags (one per input token), space-separated, no punctuation, no explanations.\newline
        \texttt{\{\{sentence content, \textbf{s}\}\}  }      
    }}}
    \end{tabular}
\end{table}

\subsection{Experimental Settings}
The subsequent baseline experiments were tested using a 40GB A100 GPU. Batch sizes were tested and set at 4 for each run for maximum accuracy and consistency. Evaluation and logging steps were also set to reflect optimization: 10 for the much smaller \textit{definite} IFL human evaluated model(s) and 1000 for the large \textit{potential} IFL dataset. All models were cross evaluated on all datasets to test generalizability.

\subsection{Evaluation Metrics}
The metrics, such as those shown in Table \ref{tab:current_sota}, are commonly used for evaluation of language models. Precision and recall help researchers use the true positives, false positives and false negatives to measure the effectiveness of models on given data and task. The F1 score formulates the combination of the precision and recall measurements, thereby giving an overall score for a given model. Sequence accuracy remains an equally important metric for inclusion, since idioms and figurative language are more often multi-word expressions.
\paragraph{Sequence Accuracy} 
Sequence accuracy is calculated by for slot labeling as the average of sequences, $s_i$, with all labels correctly assigned. For $N$ number of labels The simple formula is calculated as:

\newcommand{\I}{\mathbf{1}}

\begin{align}
\mathrm{Sequence\_Accuracy}
= \frac{1}{N}\sum_{i=1}^{N} \I\{\hat{\mathbf{s}}_i = \mathbf{s}_i\}
\end{align}
\paragraph{Entity-level performance}  
Precision, recall and F1 scores are basic scoring metrics for language models. 

\newcommand{\Egoldi}{E_i}
\newcommand{\Epredi}{\hat{E}_i}

\newcommand{\TP}{\mathrm{TP}}
\newcommand{\FP}{\mathrm{FP}}
\newcommand{\FN}{\mathrm{FN}}

\begin{align}
\TP &= \sum_{i=1}^{N} \bigl|\, \Egoldi \cap \Epredi \,\bigr|, \\
\FP &= \sum_{i=1}^{N} \bigl|\, \Epredi \setminus \Egoldi \,\bigr|, \\
\FN &= \sum_{i=1}^{N} \bigl|\, \Egoldi \setminus \Epredi \,\bigr|.
\end{align}
The precision measurement focuses on the relationship between all positive results predictions. All relevant instances are divided by all retrieved instances. For \textbf{TP}, true positives and \textbf{FP}, false positives:
\begin{align}
\mathrm{Precision}\ (P) &= \frac{\TP}{\TP + \FP}. 
\end{align}
Recall emphasizes only relevant instances by replacing false positives with \textbf{FN}, false negatives. It is measured by the following:
\begin{align}
    \mathrm{Recall}\ (R)    &= \frac{\TP}{\TP + \FN}. 
\end{align}
While recall and precision give empirical impressions of model performance on specific data, the F1 score shows a model's general ability on the data. It is formulated with precision, \textbf{P}, and recall, \textbf{R}, as follows:
\begin{align}
    \mathrm{F1}             &= \frac{2PR}{P + R}.
\end{align}

\subsection{Results and Analysis}

\begin{figure}[htbp]
    \centering
    \includegraphics[width=1\linewidth]{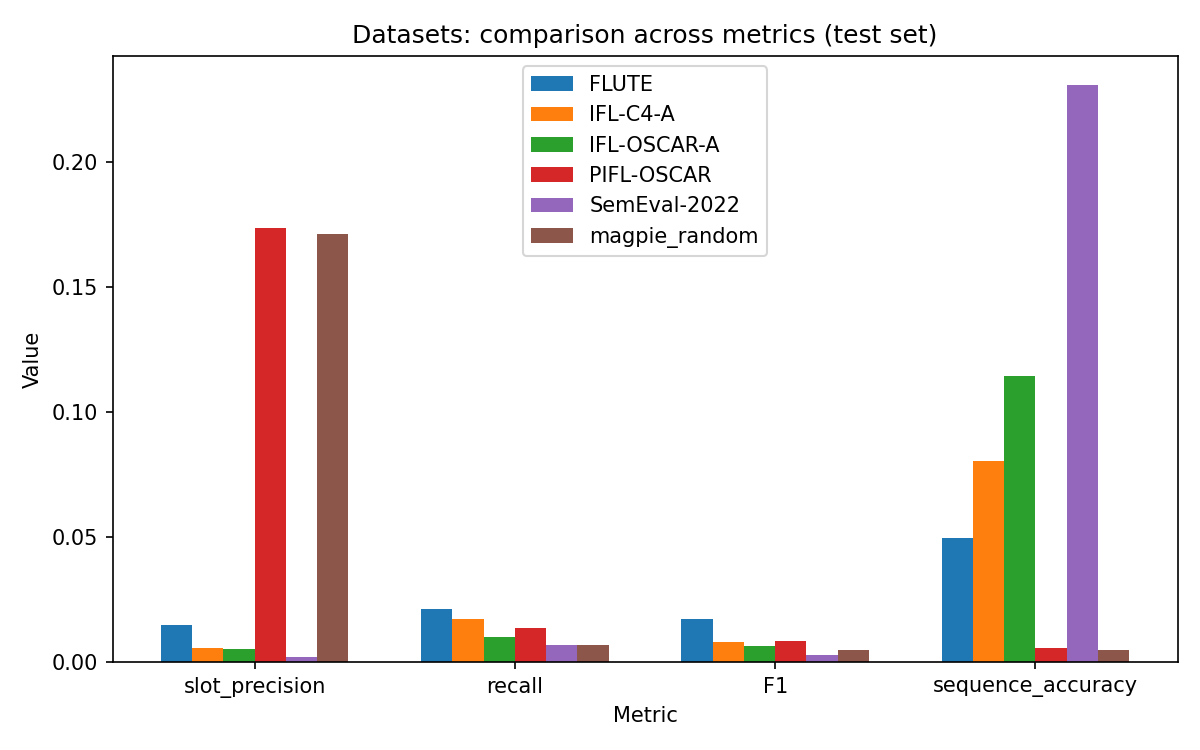}
    \caption{Average performance of all LLMs by metric and dataset.}
    \label{fig:LLM_avg}
\end{figure}

To establish additional necessity of this research and accurate baselines, the following open-weight chat LLMs were tested: Gemma-3, GPT-OSS, Llama3.1, Mistral, Phi-4, and Qwen2.5. The models were accessed and prompted via Ollama. Regardless of model size, the standard quantization on this platform is 4-bit, which helps the models perform more quickly, but decreases accuracy. Models were evaluated on slot precision, recall, F1, and sequence accuracy. The results for each test shown in Tables \ref{tab:test_slot_precision},\ref{tab:test_recall}, \ref{tab:test_f1} were generally poor except the sequence accuracy results in \ref{tab: llm_seq_acc} for the SemEval 2022 dataset. All results for chat LLMs, automatically quantized, were averaged by metric and separated by dataset in \ref{fig:LLM_avg}.

\begin{table}[htbp] 
\centering
\caption{Results for test\_slot\_precision}
\label{tab:test_slot_precision}
\small
\begin{tabular}{lrrrrrr}
\hline
Dataset & Gemma3 & GPT-OSS & Llama3.1 & Mistral & Phi4 & Qwen2.5 \\
\hline
IFL-C4-A& 0.01 & 0 & 0 & 0.01 & 0.01 & 0 \\
FLUTE& 0.02 & 0 & 0 & 0.01 & 0.04 & 0.02 \\
PIFL-OSCARPIE& 0.01 & 1 & 0 & 0.01 & 0.01 & 0.01 \\
IFL-OSCARPIE-A& 0    & 0 & 0 & 0    & 0    & 0.03 \\
SemEval-2022& 0    & 0 & 0 & 0    & 0.01 & 0 \\
magpie\_random& 0    & 1 & 0 & 0    & 0.02 & 0.01 \\
\hline
\end{tabular}
\end{table}

\begin{table}[htbp]
\centering
\caption{Results for test\_recall}
\label{tab:test_recall}
\small
\begin{tabular}{lrrrrrr}
\hline
Dataset & Gemma3 & GPT-OSS & Llama3.1 & Mistral & Phi4 & Qwen2.5 \\
\hline
IFL-C4-A
& 0.04 & 0 & 0 & 0.04 & 0.02 & 0 \\
FLUTE
& 0.03 & 0 & 0 & 0.02 & 0.05 & 0.03 \\
PIFL-OSCARPIE
& 0.03 & 0 & 0 & 0.02 & 0.01 & 0.01 \\
IFL-OSCARPIE-A
& 0.02 & 0 & 0 & 0    & 0    & 0.04 \\
SemEval-2022
& 0.02 & 0 & 0 & 0.01 & 0.02 & 0 \\
magpie\_random& 0.01 & 0 & 0 & 0    & 0.02 & 0.01 \\
\hline
\end{tabular}
\end{table}

\begin{table}[htbp]
\centering
\caption{Results for test\_f1}
\label{tab:test_f1}
\small
\begin{tabular}{lrrrrrr}
\hline
Dataset & Gemma3 & GPT-OSS & Llama3.1 & Mistral & Phi4 & Qwen2.5 \\
\hline
IFL-C4-A
& 0.01 & 0 & 0 & 0.02 & 0.02 & 0 \\
FLUTE
& 0.02 & 0 & 0 & 0.01 & 0.04 & 0.03 \\
PIFL-OSCARPIE
& 0.02 & 0 & 0 & 0.01 & 0.01 & 0.01 \\
IFL-OSCARPIE-A
& 0.01 & 0 & 0 & 0    & 0    & 0.03 \\
SemEval-2022
& 0    & 0 & 0 & 0    & 0.01 & 0 \\
magpie\_random& 0.01 & 0 & 0 & 0    & 0.02 & 0.01 \\
\hline
\end{tabular}
\end{table}

\begin{table}[htbp]
\centering
\caption{Results for test\_sequence\_accuracy}
\label{tab:test_sequence_accuracy}
\small
\begin{tabular}{lrrrrrr}
\hline
Dataset & Gemma3 & GPT-OSS & Llama3.1 & Mistral & Phi4 & Qwen2.5 \\
\hline
IFL-C4-A
& 0.02 & 0.2  & 0.2  & 0.03 & 0    & 0.03 \\
FLUTE
& 0.03 & 0.09 & 0.09 & 0.02 & 0.03 & 0.02 \\
PIFL-OSCARPIE
& 0    & 0    & 0    & 0    & 0.01 & 0.01 \\
IFL-OSCARPIE-A
& 0.06 & 0.3  & 0.3  & 0.03 & 0    & 0 \\
SemEval-2022
& 0.16 & 0.45 & 0.44 & 0.05 & 0.11 & 0.18 \\
magpie\_random& 0    & 0    & 0    & 0    & 0.01 & 0 \\
\hline
\end{tabular}
\label{tab: llm_seq_acc}
\end{table}

\subsubsection{LLM errors}
Error analysis of the chat LLMs on a one-shot prompt setting show highly inaccurate predictions. A common error for each model within each dataset was chosen for analysis. Tables \ref{tab: LLM_errors_1}, \ref{tab: LLM_errors_2} show a variety of incorrect labeling schemes. The types include no labels, incomplete target phrase labeling, incorrect phrase labeling: \textit{shipshape B-Idiom[and] I-Idiom[bristol]} (FLUTE), partial target phrase label combined with non-target phrase label: \textit{get in B-Idiom[touch] I-Idiom[if]} (IFL-C4-A), and over-labeling the phrase with additional words: \textit{B-idiom[lead] I-idiom[the] I-idiom[way] I-idiom[by] I-idiom[trying]} (\textit{P}IFL-OSCAR).  One notable exception within the error choices occurs: Mistral correctly predicts \textit{all shipshape} in the FLUTE dataset. 
\begin{table*}[htbp]
    \centering
    \caption{Examples of errors from chat LLMs for IFL-C4-A, FLUTE, \textit{P}IFL-OSCAR, and IFL-OSCAR-A datasets.}
    \label{tab: LLM_errors_1}
    \resizebox{0.99\linewidth}{!}{
    \begin{tabular}{p{0.23\linewidth}p{0.03\linewidth}p{0.94\linewidth}}
        \hline
        \textbf{Dataset} & & \textbf{Sentence and Model Outputs} \\
        \hline

        \textbf{IFL-C4-A} & & \\
        \textit{Sentence} & & Please get in touch if you would like to discuss your grant idea. \\
        \textit{Gold Labels} & & Please B-idiom[get] I-idiom[in] I-idiom[touch] if you would like to discuss your grant idea. \\
        \textit{Gemma3} & \xmark & Please B-idiom[get] I-idiom[in] I-idiom[touch] I-idiom[if] I-idiom[you] I-idiom[would] I-idiom[like] to discuss your grant idea. \\
        \textit{GPT-OSS} & \xmark & Please get in touch if you would like to discuss your grant idea. \\
        \textit{Llama3.1} & \xmark & Please get in touch if you would like to discuss your grant idea. \\
        \textit{Mistral} & \xmark & Please get in touch if you would like to discuss your grant idea. \\
        \textit{Phi4} & \xmark & Please get in B-idiom[touch] I-idiom[if] you would like to discuss your grant idea. \\
        \textit{Qwen2.5} & \xmark & Please B-idiom[get] I-idiom[in] touch if you would like to discuss your grant idea. \\
        \hline

        \textbf{FLUTE} \\
        \textit{Sentence} & & All shipshape and bristol fashion. \\
        \textit{Gold Labels} & & B-idiom[All] I-idiom[shipshape] and bristol fashion. \\
        \textit{Gemma3} & \xmark & All shipshape and bristol fashion. \\
        \textit{GPT-OSS} & \xmark & All shipshape and bristol fashion. \\
        \textit{Llama3.1} & \xmark & All shipshape and bristol fashion. \\
        \textit{Mistral} & \checkmark & B-idiom[All] I-idiom[shipshape] and bristol fashion. \\
        \textit{Phi4} & \xmark & All shipshape B-idiom[and] I-idiom[bristol] I-idiom[fashion] . \\
        \textit{Qwen2.5} & \xmark & All shipshape B-idiom[and] I-idiom[bristol] fashion. \\
        \hline

        \textbf{\textit{P}IFL-OSCAR} & & \\
        \textit{Sentence} & & ``California can lead the way by trying to bring them home.'' \\
        \textit{Gold Labels} & & California can B-Idiom[lead] I-idiom[the] I-idiom[way] by trying to bring them home. \\
        \textit{Gemma3} & \xmark & B-idiom[California] I-idiom[can] I-idiom[lead] I-idiom[the] I-idiom[way] I-idiom[by] I-idiom[trying] to bring them home. \\
        \textit{GPT-OSS} & \xmark & California can lead the way by trying to bring them home. \\
        \textit{Llama3.1} & \xmark & California can lead the way by trying to bring them home. \\
        \textit{Mistral} & \xmark & California can lead the way by trying to bring them home. \\
        \textit{Phi4} & \xmark & California can B-idiom[lead] I-idiom[the] I-idiom[way] I-idiom[by] I-idiom[trying] to bring them home. \\
        \textit{Qwen2.5} & \xmark & B-idiom[California] can lead the way by trying to bring them home. \\
        \hline

        \textbf{IFL-OSCAR-A} & & \\
        \textit{Sentence} & & Everyone wants a chance to spill their guts now and then. \\
        \textit{Gold Labels} & & Everyone wants a chance to B-idiom[spill] I-idiom[their] I-idiom[guts] now and then. \\
        \textit{Gemma3} & \xmark & B-idiom[Everyone] I-idiom[wants] I-idiom[a] I-idiom[chance] I-idiom[to] I-idiom[spill] I-idiom[their] guts now and then. \\
        \textit{GPT-OSS} & \xmark & Everyone wants a chance to spill their guts now and then. \\
        \textit{Llama3.1} & \xmark & Everyone wants a chance to spill their guts now and then. \\
        \textit{Mistral} & \xmark & Everyone wants a chance to spill their guts now and then. \\
        \textit{Phi4} & \xmark & Everyone wants a B-idiom[chance] I-idiom[to] I-idiom[spill] I-idiom[their] I-idiom[guts] now and then. \\
        \textit{Qwen2.5} & \xmark & Everyone wants B-idiom[a] I-idiom[chance] to spill their guts now and then. \\
        \hline

    \end{tabular}
    }
  
\end{table*}

\begin{table*}[htbp]
    \centering
    \caption{Examples of errors from chat LLMs for SemEval-2022 and magpie\_random datasets.}
    \label{tab: LLM_errors_2}
    \resizebox{0.99\linewidth}{!}{
    \begin{tabular}{p{0.23\linewidth}p{0.03\linewidth}p{0.94\linewidth}}
        \hline
        \textbf{Dataset} & & \textbf{Sentence and Model Outputs} \\
        \hline
        \textbf{SemEval 2022} & & \\
        \textit{Sentence} & & Somehow the Steelers had jitters for this game. And somehow the Steelers laid the biggest goose egg of a game plan of the year. This was as bad as it has ever as been in the Tomlin era.\\
        \textit{Gold Labels} & & Somehow the Steelers had jitters for this game. And somehow the Steelers laid the biggest B-Idiom[goose] I-Idiom[egg] of a game plan of the year. This was as bad as it has ever as been in the Tomlin era.\\
        \textit{Gemma3} & \xmark &  B-idiom[Somehow]  B-idiom[the] Steelers  I-idiom[had]  B-idiom[jitters]  I-idiom[for] this  I-idiom[game].  B-idiom[And]  I-idiom[somehow] the  I-idiom[Steelers]  B-idiom[laid]  I-idiom[the] biggest  I-idiom[goose]  B-idiom[egg]  I-idiom[of] a game plan of the year. This was as bad as it has ever as been in the Tomlin era.\\
        \textit{GPT-OSS} & \xmark & Somehow the Steelers had jitters for this game. And somehow the Steelers laid the biggest goose egg of a game plan of the year. This was as bad as it has ever as been in the Tomlin era. \\
        \textit{Llama3.1} & \xmark & Somehow the Steelers had jitters for this game. And somehow the Steelers laid the biggest goose egg of a game plan of the year. This was as bad as it has ever as been in the Tomlin era.\\
        \textit{Mistral} & \xmark &  B-idiom[Somehow] the Steelers  I-idiom[had] jitters  B-idiom[for] this  B-idiom[game]. And  I-idiom[somehow] the  B-idiom[Steelers] laid the biggest goose egg of a game plan of the year. This was as bad as it has ever as been in the Tomlin era. \\
        \textit{Phi4} & \xmark & Somehow the Steelers had  B-idiom[jitters]  I-idiom[for] this game. And somehow the Steelers laid the biggest goose egg of a game plan of the year. This was as bad as it has ever as been in the Tomlin era.\\
        \textit{Qwen2.5} & \xmark & Somehow  B-idiom[the]  I-idiom[Steelers] had jitters for this game. And somehow the Steelers laid the biggest goose egg of a game plan of the year. This was as bad as it has ever as been in the Tomlin era.\\
        \hline

        \textbf{magpie\_random} & & \\
        \textit{Sentence} & & ``Leith , while her voice stayed calm , jumped the gun .'' \\
        \textit{Gold Labels} & & Leith , while her voice stayed calm , B-Idiom[jumped] I-Idiom[the] I-Idiom[gun] . \\
        \textit{Gemma3} & \xmark & B-idiom[Leith] I-idiom[,] while I-idiom[her] voice I-idiom[stayed] calm B-idiom[,] I-idiom[jumped] the gun .
      \\
        \textit{GPT-OSS} & \xmark & Leith , while her voice stayed calm , jumped the gun .
       \\
        \textit{Llama3.1} & \xmark & Leith , while her voice stayed calm , jumped the gun . 
        \\
        \textit{Mistral} & \xmark & B-idiom[Leith] I-idiom[,] while her voice stayed I-idiom[calm] , jumped the gun . 
        \\
        \textit{Phi4} & \xmark &Leith , while B-idiom[her] I-idiom[voice] I-idiom[stayed] calm , jumped the gun . 
      \\
        \textit{Qwen2.5} & \xmark & Leith , B-idiom[while] I-idiom[her] voice stayed calm , jumped the gun .
        O O B-idiom I-idiom O O O O O O O O\\
        \hline
    \end{tabular}
    }
    
\end{table*}

For baseline testing of the slot loss architecture, both BERT and RoBERTa were used. These were finetuned on each dataset with a training/development/test split of 80/10/10 percent, respectively. They were subsequently optimized on the development set. Evaluation occurred on the remaining 10 percent test set. The first positive indication for any model are the test set results compared with the development set results. Evaluation on the test set for each model were consistently higher, while BERT-trained models, rather than RoBERTa-trained models performed better for all metrics, as seen in Table \ref{tab:main_result_on_our_datasets}. 
\begin{table*}[htbp]
\centering
\caption{Overall result of Bert-based models on our human-annotated datasets.} 
\label{tab:main_result_on_our_datasets}
\resizebox{\linewidth}{!}{%
\begin{tabular}{l c c c c l c}
    \toprule
    \textbf{Dataset} & \textbf{F1} & \textbf{Recall} & \textbf{Sequence Acc.}& \textbf{Slot precision} & \textbf{Model}& \textbf{Batch size}\\\midrule
    IFL-OSCAR-A& .7708 & .8222 & .8000 & .7255 & BERT & 4 \\
    IFL-OSCAR-A& .6882 & .7111 & .7286 & .6596 & RoBERTa & 4 \\
    \textit{P}IFL-C4-A& .8913 & .9318 & .8833 & .8542 & BERT & 4 \\
    \textit{P}IFL-C4-A& .8542 & .9318 & .8167 & .7885 & RoBERTa & 4 \\
    \textit{P}IFL-OSCAR & .9924 & .9947 & .9893 & .9902 & BERT & 4 \\
    \textit{P}IFL-OSCAR & .9895 & .9922 & .9857 & .9868 & RoBERTa & 4 \\
    \bottomrule
\end{tabular}
}
\end{table*}

When this baseline architecture was applied to the most recent existing dataset, magpie\_random, we received competitive results for precision and F1. The results for sequence accuracy achieved state of the art, observed in Fig. \ref{tab:main_result_on_magpie}, compared to the results received by cohesion loss and retranslation \cite{yayavaram-etal-2024-bert} at 0.9151.

\begin{table*}[htbp]
\centering
\caption{Overall results of Bert-based models on magpie\_random. SOTA achieved for Sequence accuracy in all settings.} 
\label{tab:main_result_on_magpie}
\resizebox{\linewidth}{!}{%
\begin{tabular}{l c c c c }
    \toprule
    \textbf{Dataset} & \textbf{F1} & \textbf{Recall} & \textbf{Sequence Acc.}& \textbf{Slot precision}\\\midrule
    BERT test& .9313 & .9436 & \textbf{.9297} & .9193 \\
    BERT dev & .9273 & .9403 & \textbf{.9253} & .9147  \\
    RoBERTa test & .9289 & .9376 & \textbf{.9254} & .9205 \\
    RoBERTa dev & .9270 & .9362 & \textbf{.9226} & .9180 \\
    
    \bottomrule
\end{tabular}
}
\end{table*}

\begin{figure}[htbp]
    \centering
    \includegraphics[width=1\linewidth]{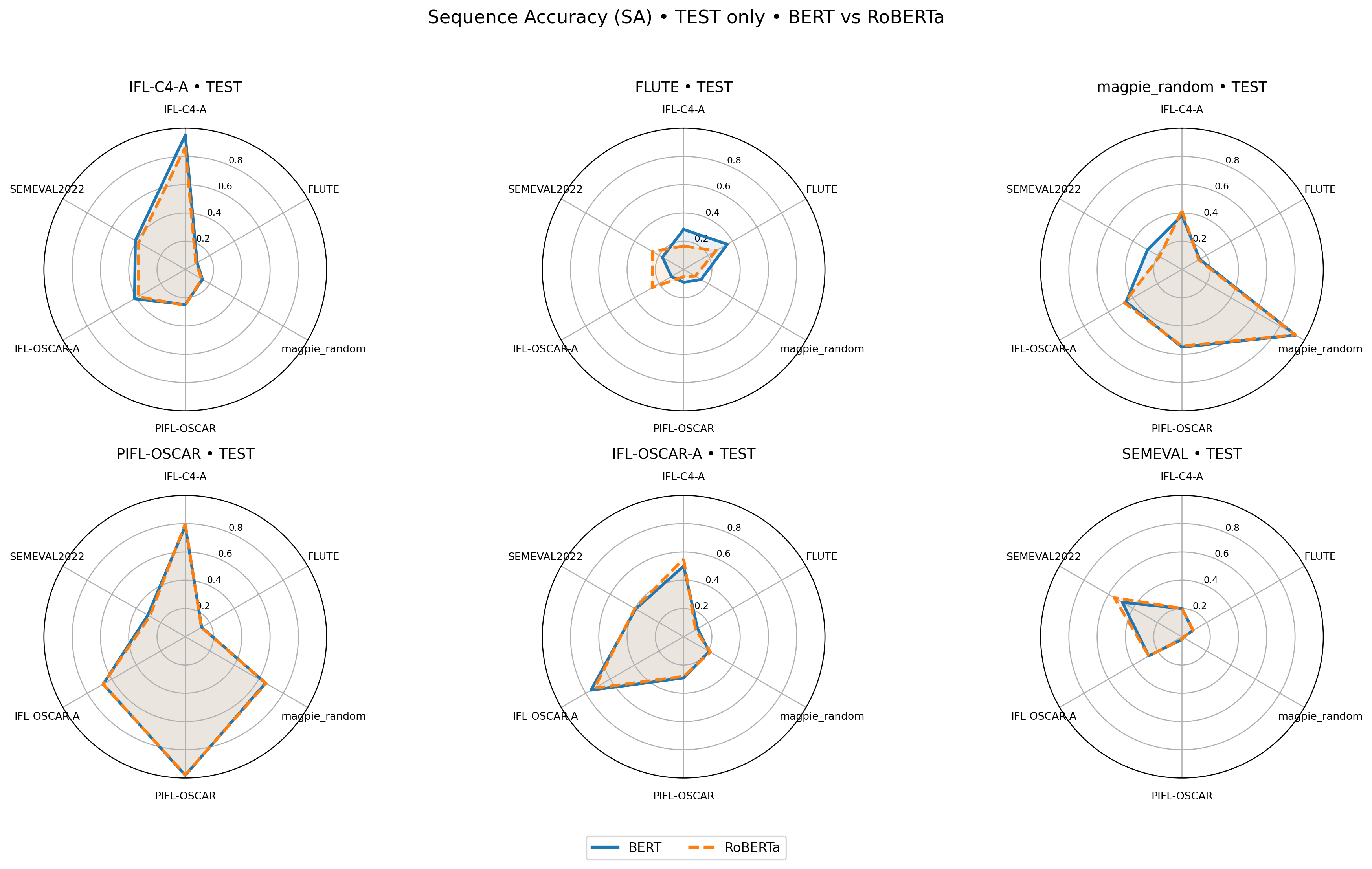}
    \caption{Cross evaluation results for sequence accuracy on test sets. BERT and RoBERTa models trained with each dataset are evaluated on the test set of all six datasets.}
    \label{fig: test_SA}
\end{figure}

\begin{figure}[htbp]
    \centering
    \includegraphics[width=1\linewidth]{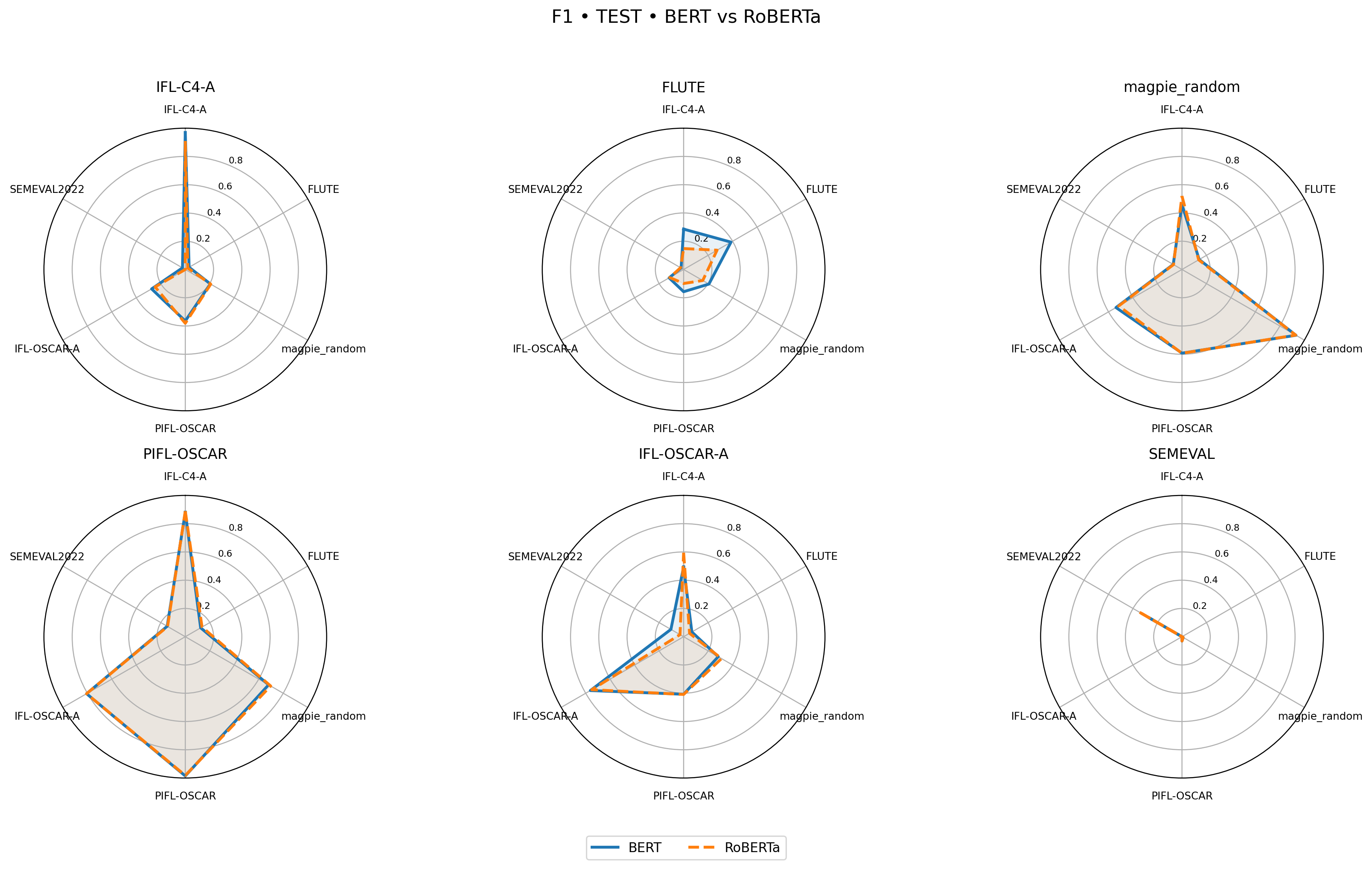}
    \caption{Cross evaluation results for F1 on test sets. BERT and RoBERTa models trained with each dataset are evaluated on the test set of all six datasets.}
    \label{fig: test_F1}
\end{figure}

Each dataset was initially trained and self-evaluated on the same dataset's test split. The models clearly performed best on this setting, but to establish generalizability of each model, a cross evaluation was performed.  BERT- and RoBERTa-based models were trained on each dataset, but evaluated on the test sets for the remaining datasets. Development set results were also acquired. Additional results of development data in Figures \ref{fig:dev_f1}, \ref{fig:dev_sa} and tables of raw data  are found in the appendix, \ref{table:raw_dev}.
The generalization performance is directly related to the area visualized for sequence accuracy in Fig. \ref{fig: test_SA} and F1 in Fig. \ref{fig: test_F1}. The data revealed that the human annotated data did not evaluate well on other datasets, while the models trained on the \textit{P}IFL-OSCAR dataset tested well and generalized the best. According to generalizability, \textit{P}IFL-OSCAR is followed by models trained on magpie{\_}random, IFL-OSCAR-A, IFL-C4-A, with FLUTE and SemEval-2022-trained models generalizing the least.

An interesting subset of sentences within the dataset consists exclusively of idiomatic expressions. The phrase \textit{“Piece of cake!”} exemplifies this category. Such sentences inherently lack surrounding contextual information, resulting in minimal or non-existent intra-sentence cosine similarity data. Consequently, the inclusion of adjacent sentences—both preceding and following—is crucial to capture the inter-sentence cosine similarities that provide broader semantic grounding. By incorporating neighboring context, it becomes possible to more accurately evaluate figurative usage and maintain a coherent representation of semantic relationships, particularly when idioms occur in isolation.

A deeper examination of the errors contributing to these outcomes reveals patterns in the labeling performance, as summarized in Tables, \ref{tab: LLM_errors_1}, \ref{tab: LLM_errors_2},  \ref{tab: ifl-errors}, which report BIO tag discrepancies for both the BERT-based models and the chat-oriented large language models (LLMs). During cross-evaluation, a slight reduction was observed between the precision and recall values of models trained on the \textit{P}IFL-OSCAR dataset. A widening gap between these metrics can be indicative of over-generalization, a phenomenon influenced by both the architecture of the sequence tagging framework and the methodological emphasis applied during data acquisition. False negatives, as seen in Table \ref{tab:false_neg}, also show near and complete misses by the baseline models. Near misses include an \textit{I-Idiom} label in the first position where \textit{B-Idiom} is the logical target at the beginning of a figurative expression. A complete miss includes no label for a labeled word or vice versa.

\begin{table*}[htbp]
            \centering 

           \caption{Examples of errors from annotated datasets IFL-C4-A and IFL-OSCAR-A.}
           \label{tab: ifl-errors}
            \resizebox{\columnwidth}{!}{%
            \begin{tabular}{p{0.28\textwidth}cp{0.82\textwidth}}
                \toprule
                Errors & & \\
                \midrule
                    \textit{Sentence} && It has a smooth , comfortable ride with room for five and generous cargo space \textbf{to boot}.\\
                    \textit{Gold Labels}&& It has a smooth , comfortable ride with room for five and generous cargo space B-idiom[to] I-idiom[boot]. \\
                    \textit{IFL-C4-A-BERT}&\xmark&  It has a smooth , comfortable ride with room for five and generous cargo space to boot.
\\
 \textit{IFL-C4-A-RoBERTa}& \xmark&It has a smooth , comfortable ride with room for five and generous cargo space to boot.
\\
 \midrule
 \textit{Sentence} & &Center parts are great for sleek , \textbf{squeaky clean} locks .\\
 \textit{Gold Labels}& &Center parts are great for sleek , B-idiom[squeaky] I-idiom[clean] locks .
\\
 \textit{IFL-C4-A-BERT}
& \xmark&Center parts are great for sleek , B-idiom[squeaky] clean locks .
\\ 
                     \textit{IFL-C4-A-RoBERTa}&\xmark& Center parts are great for sleek , B-idiom[squeaky] clean locks .
\\
\toprule 
 Correct Predictions& &\\
 \midrule
 \textit{Sentence} 
& &How do we seek meaning where none is to be found - - and can we create it \textbf{from scratch} ?\\
 \textit{Gold Labels}& &How do we seek meaning where none is to be found - - and can we create it B-Idiom[from] I-Idiom[scratch] ?\\
 \textit{IFL-C4-A-BERT}
& \checkmark&How do we seek meaning where none is to be found - - and can we create it B-Idiom[from] I-Idiom[scratch] ?\\
 \textit{IFL-C4-A-RoBERTa}& \checkmark&How do we seek meaning where none is to be found - - and can we create it B-Idiom[from] I-Idiom[scratch] ?\\
\toprule
                Errors & & \\
                \midrule
                    \textit{Sentence} 
&&And anyway , isn ' t this a fairly horrible thing to \textbf{bank on} \\ 
                    \textit{Gold Labels}
&& And anyway , isn ' t this a fairly horrible thing to B-idiom[bank] I-idiom[on] ?
\\ 
                    \textit{IFL-OSCAR-A-BERT}
&\xmark& And anyway , isn ' t this a fairly horrible thing to B-idiom[bank] on ?
\\
                    \textit{IFL-OSCAR-A-RoBERTa}&\xmark& And anyway , isn ' t this a fairly horrible thing to bank on ?
\\
\midrule 
 \textit{Sentence} & &Learn more \textbf{The bottom line is} we value our customers , and take the additional steps needed in production , research , development and support to provide the finest product and service available .\\
 \textit{Gold Labels}& &Learn more B-idiom[The] I-idiom[bottom] I-idiom[line] I-idiom[is] we value our customers , and take the additional steps needed in production , research , development and support to provide the finest product and service available .\\ 
            
                    \textit{IFL-OSCAR-A-BERT}
&\xmark& Learn more I-idiom[The] I-idiom[bottom] I-idiom[line] I-idiom[is] we value our customers , and take the additional steps needed in production , research , development and support to provide the finest product and service available .
\\ 
                    \textit{IFL-OSCAR-A-RoBERTa}&\xmark& Learn more B-idiom[The] I-idiom[bottom] I-idiom[line] is we value our customers , and take the additional steps needed in production , research , development and support to provide the finest product and service available .\\
\toprule 
 Correct Predictions& &\\
 \midrule
 \textit{Sentence} 
& &This is a smaller office , so the wait times are \textbf{next to nothing} !\\
 \textit{Gold Labels}
& &This is a smaller office , so the wait times are B-idiom[next] I-idiom[to] I-idiom[nothing] !\\
 \textit{IFL-OSCAR-A-BERT}
& \checkmark&This is a smaller office , so the wait times are B-idiom[next] I-idiom[to] I-idiom[nothing] !\\
 
\textit{IFL-OSCAR-A-RoBERTa}& \checkmark&This is a smaller office , so the wait times are B-idiom[next] I-idiom[to] I-idiom[nothing] !\\
\bottomrule
            \end{tabular}
            }
        \end{table*} 

\begin{table}[htbp]
    \centering
    \caption{False negatives observed in error analysis. The target sentence contained other figurative language, extended figurative expression, or the model falsely predicted without a beginning idiom word.}
    \label{tab:false_neg}
    \begin{tabular}{p{0.24\linewidth} p{0.70\linewidth}}
        \toprule
        \multicolumn{2}{l}{\textbf{FALSE NEGATIVES}} \\
        \midrule
        \textit{Gold Labels} & When Aurora first came to London in 2016 she never thought that the songs she had collected over the years would ever see I-idiom[the] I-idiom[light] of day. \\
        \textit{IFL-OSCAR-A-BERT} & When Aurora first came to London in 2016 she never thought that the songs she had collected over the years would ever see I-idiom[the] I-idiom[light] I-idiom[of] I-idiom[day]. \\
        \textit{IFL-OSCAR-A-RoBERTa} & When Aurora first came to London in 2016 she never thought that the songs she had collected over the years would ever see I-idiom[the] I-idiom[light] I-idiom[of] I-idiom[day]. \\
        \midrule
        \textit{Gold Labels} & {company name} On the same note, platform fees, paid trading indicators and other trading software can also be an issue and you should evaluate your need for the trading platform that has all the B-idiom[bells] I-idiom[and] I-idiom[whistles]. \\
        \textit{IFL-OSCAR-A-BERT} & {company name} B-idiom[On] I-idiom[the] I-idiom[same] I-idiom[note], platform fees, paid trading indicators and other trading software can also be an issue and you should evaluate your need for the trading platform that has all the B-idiom[bells] I-idiom[and] I-idiom[whistles]. \\
        \textit{IFL-OSCAR-A-RoBERTa} & Correct according to gold. \\
        \midrule
        \textit{Gold Labels} & [username]'s instructions are not only in Dutch, but pretty bare bones, so I didn't bother translating them.\\
        \textit{IFL-OSCAR-A-BERT} & (no prediction) \\
        \textit{IFL-OSCAR-A-RoBERTa} & [username]'s instructions are not only in Dutch, but pretty B-idiom[bare] I-idiom[bones], so I didn't bother translating them.\\
        \bottomrule
    \end{tabular}
\end{table}

The annotation of figurative expressions through direct human evaluation was deemed necessary, since figurativeness was inferred when identical or closely matching tokens appeared in the construction of the larger dataset. While the approach facilitated consistent labeling, it also yielded spurious false positives during self-evaluation due to the model’s tendency to rely on surface-level lexical features rather than deeper contextual signals. Though, when assessed under cross-evaluation, the generalization patterns remain stable for \textit{P}IFL-OSCAR, as seen in both sequence accuracy and F1 score Figs. \ref{fig: test_SA},\ref{fig: test_F1}. 


\section{Conclusion}

A large-scale, targeted dataset of colloquial language was constructed from Creative Commons sources to capture the extensive range of \textit{potential} idiomatic expressions present in natural discourse. Afterwards, a derivative dataset was retrieved from the C4 corpus following equivalent collection and filtering methods. Additional refinement was achieved through human evaluation of both \textit{P}IFL-OSCAR and \textit{P}IFL-C4, resulting in the creation of the IFL-OSCAR-A and IFL-C4-A annotated subsets. These curated datasets collectively contribute to a collection of training-agnostic datasets that explicitly distinguish between idiomatic and non-idiomatic instances of target expressions, and thereby present new opportunities and challenges for consistent evaluation across model architectures and linguistic domains.

Despite these advancements, several limitations merit consideration. The data sanitization process, the presence of possibly machine-generated text in the 2018 snapshot of Common Crawl, and the lexical matching of idiomatic expressions inherently constrain the larger dataset to \textit{possible} rather than \textit{definite} idioms. When combined with cosine similarity measures and linguistic priors—such as syntactic position and part-of-speech distribution—a probabilistic threshold for figurativeness can be estimated and used in hyperparameter optimization.

Future iterations of this dataset could incorporate a confidence metric representing the probability of idiomaticity for each instance. Such an adaptation would facilitate more nuanced model evaluation and training, particularly in determining figurative language type. Extending the dataset to include multilingual variants would enhance its generalizability across languages and enable broader cross-linguistic analyses of idiomatic expressions within natural language understanding frameworks.

\appendix
\begin{landscape}
\begingroup
\thispagestyle{empty}
\makeatletter
\@ifpackageloaded{hyperref}{\hypersetup{pageanchor=false}}{}
\makeatother
\let\origthepage\thepage
\renewcommand{\thepage}{}

\refstepcounter{section}%
\section*{\thesection\quad Tables of Raw Cross Evaluation Values}%
\addcontentsline{toc}{section}{Tables of Raw Cross Evaluation Values}%
\label{app:cross-eval}

\setlength{\tabcolsep}{3.5pt}
\renewcommand{\arraystretch}{1.0}
\small

\begin{center}
\label{table:raw_test}
\captionof{table}{Cross evaluation results on BERT-based models (Test set). For each column, each metric is shown. (Sequence Accuracy/F1/Precision/Recall)}
\label{tab_cross_eval_test}
\resizebox{0.98\linewidth}{!}{%
\begin{tabular}{|l|c|c|c|c|c|c|}
\hline
\textbf{Dataset} & \textbf{IFL-C4-A} & \textbf{FLUTE} & \textbf{magpie-random} & \textbf{PIFL-OSCAR} & \textbf{IFL-OSCAR-A} & \textbf{SemEval-2022} \\
\hline
IFL-C4-A-BERT & 95/96.91/95/97.2 & 9.68/3.12/13.95/1.75 & 14.02/20.44/33.2/14.76 & 24.76/36.46/57.83/26.62 & 41.43/27.5/36.67/22 & 40.81/2.29/5.76/1.43 \\
\hline
IFL-C4-A-RoBERTa & 86.67/90.91/88.24/93.75 & 8.62/1.99/6.67/1.17 & 13.44/20.78/38.23/14.27 & 25.42/38.08/65.08/27.77 & 38.57/24.66/39.13/18 & 38.06/0/0/0 \\
\hline
FLUTE-BERT & 28.33/28.57/26.32/31.25 & 35.68/38.7/37.55/39.91 & 14.24/20.67/19.22/22.35 & 9.06/15.74/14.89/16.69 & 10/11.85/9.41/16 & 17.32/2.06/1.55/3.09 \\
\hline
FLUTE-RoBERTa & 16.67/14.81/18.18/12.5 & 26.53/27.31/28.55/26.17 & 9.46/15.62/20.75/12.53 & 5.08/9.92/13.14/7.97 & 25.71/11.36/13.16/10 & 25.33/2.49/2.39/2.62 \\
\hline
magpie-random-BERT & 38.33/45.99/44.23/47.92 & 14.46/14.07/14.41/13.74 & 92.97/93.13/91.93/94.36 & 54.95/59.21/61.89/56.76 & 45.71/53.91/47.69/62 & 28.08/7.14/7.71/6.65 \\
\hline
magpie-random-RoBERTa & 41.67/52/50/54.17 & 13.66/13.64/13.83/13.45 & 92.54/92.89/92.05/93.76 & 54.18/59.31/63.21/55.87 & 47.14/51.67/44.29/62 & 18.37/6.89/5.88/8.33 \\
\hline
PIFL-OSCAR-BERT & 78.33/88.07/78.69/100 & 13.39/12.56/13.56/11.69 & 65.54/68.07/66.93/69.26 & 98.06/98.55/98.05/99.04 & 67.14/80.65/67.57/100 & 30.58/14.79/15.09/14.49 \\
\hline
PIFL-OSCAR-RoBERTa & 80/88.89/80/100 & 13.13/13.67/13.89/13.45 & 66.03/69.75/70.63/68.89 & 97.71/98.19/97.79/98.59 & 67.14/80.65/67.57/100 & 29/14.47/14.42/14.52 \\
\hline
IFL-OSCAR-A-BERT & 20/0/0/0 & 9.02/0/0/0 & 1.01/1.03/10.21/0.54 & 1.92/3.11/22.89/1.67 & 27.14/0/0/0 & 48.68/27.27/35.63/22.09 \\
\hline
IFL-OSCAR-A-RoBERTa & 20/0/0/0 & 9.15/0/0/0 & 0.92/0.94/9.05/0.49 & 1.98/3.04/19.94/1.65 & 27.14/0/0/0 & 55.25/35.57/45.86/29.05 \\
\hline
SEMEVAL-BERT & 50/49.38/60.61/41.67 & 11.41/6.73/18.92/4.09 & 20.92/28.28/38.57/22.33 & 29.13/40.72/54.76/32.41 & 75.71/76.19/72.73/80 & 39.24/10.39/17.61/7.36 \\
\hline
SEMEVAL-RoBERTa & 55/59.46/84.62/45.83 & 9.81/4.86/11.6/3.07 & 21.88/30.95/47.59/22.93 & 28.05/40.85/59.52/31.09 & 72.86/75/72.22/78 & 39.89/2.96/6.61/1.9 \\
\hline
\end{tabular}
}
\end{center}

\vspace{6mm}

\begin{center}
\label{table:raw_dev}
\captionof{table}{Cross evaluation results on BERT-based models (Dev set). For each column, each metric is shown. (Sequence Accuracy/F1/Precision/Recall)}
\label{tab_cross_eval_dev}
\resizebox{0.98\linewidth}{!}{%
\begin{tabular}{|l|c|c|c|c|c|c|}
\hline
\textbf{Dataset} & \textbf{IFL-C4-A} & \textbf{FLUTE} & \textbf{magpie-random} & \textbf{PIFL-OSCAR} & \textbf{IFL-OSCAR-A} & \textbf{SemEval-2022} \\
\hline
IFL-C4-A-BERT & 88.33/89.13/85.42/93.18 & 11.55/3.46/17.33/1.92 & 14.584/21.17/33.81/15.41 & 24.38/36.03/56.03/26.55 & 61.43/53.66/59.46/48.89 & 45.87/0/0/0 \\
\hline
IFL-C4-A-RoBERTa & 81.67/85.42/78.85/93.18 & 11.95/4.76/15.57/2.81 & 14.63/22.39/41.07/15.39 & 24.91/37.42/59.85/27.22 & 50/34.86/48/26.67 & 43.57/0.41/0.81/0.28 \\
\hline
FLUTE-BERT & 26.67/23.3/20.34/27.27 & 38.38/38.98/37.41/40.68 & 13.66/20.79/19.16/22.74 & 9.39/16.17/15.42/17.01 & 21.43/23.33/18.67/31.11 & 19.08/1.39/1.01/2.21 \\
\hline
FLUTE-RoBERTa & 26.67/17.58/17.02/18.18 & 29.75/31.71/32.21/31.21 & 9.36/15.24/19.85/12.37 & 5.71/10.49/13.97/8.41 & 30/16.67/15.69/17.78 & 27.19/0.48/0.43/0.55 \\
\hline
magpie-random-BERT & 36.67/46.46/41.82/52.27 & 14.74/14.99/15.19/14.79 & 92.53/92.73/91.47/94.03 & 55.46/59.48/61.85/57.29 & 54.29/65.49/54.41/82.22 & 31.53/6.49/6.65/6.35 \\
\hline
magpie-random-RoBERTa & 41.67/53.47/47.37/61.36 & 14.08/13.74/14.19/13.31 & 92.26/92.7/91.8/93.62 & 54.24/58.83/61.94/56.02 & 48.57/61.95/51.47/77.78 & 18.67/5.35/4.36/6.91 \\
\hline
PIFL-OSCAR-BERT & 68.33/82.24/69.84/100 & 15.14/14.84/15.91/13.91 & 66.03/68.34/67.19/69.54 & 98.34/98.82/98.31/99.35 & 61.43/76.27/61.64/100 & 33.56/11.59/11.32/11.88 \\
\hline
PIFL-OSCAR-RoBERTa & 68.33/82.24/69.84/100 & 14.21/13.85/13.65/14.05 & 66.77/70.49/70.82/70.17 & 97.88/98.45/97.93/98.98 & 61.43/76.52/62.86/97.78 & 28.82/8.39/8/8.84 \\
\hline
IFL-OSCAR-A-BERT & 26.67/0/0/0 & 10.22/0/0/0 & 0.88/0.94/10.05/0.49 & 1.75/2.89/21.79/1.55 & 34.29/0/0/0 & 53.72/26.77/32.17/22.93 \\
\hline
IFL-OSCAR-A-RoBERTa & 28.33/0/0/0 & 10.09/0/0/0 & 0.76/0.68/6.27/0.36 & 1.79/2.95/20.16/1.59 & 35.71/0/0/0 & 55.75/33.67/42.44/27.9 \\
\hline
SEMEVAL-BERT & 51.67/44.16/51.52/38.64 & 11.69/5.57/15.33/3.4 & 21.69/28.71/38.29/22.96 & 29.67/41.61/55.22/33.38 & 80/77.08/72.55/82.22 & 45.06/5.55/12.26/3.59 \\
\hline
SEMEVAL-RoBERTa & 63.33/66.67/80.65/56.82 & 12.35/6.88/15.31/4.44 & 22.01/31.76/47.74/23.79 & 28.22/40.91/59.42/31.19 & 72.86/68.82/66.67/71.11 & 43.44/1.26/2.65/0.83 \\
\hline
\end{tabular}
}
\end{center}
\clearpage
\renewcommand{\thepage}{\origthepage}
\makeatletter
\@ifpackageloaded{hyperref}{\hypersetup{pageanchor=true}}{}
\makeatother
\endgroup
\end{landscape}

\begin{figure}

    \centering
    \includegraphics[width=1\linewidth]{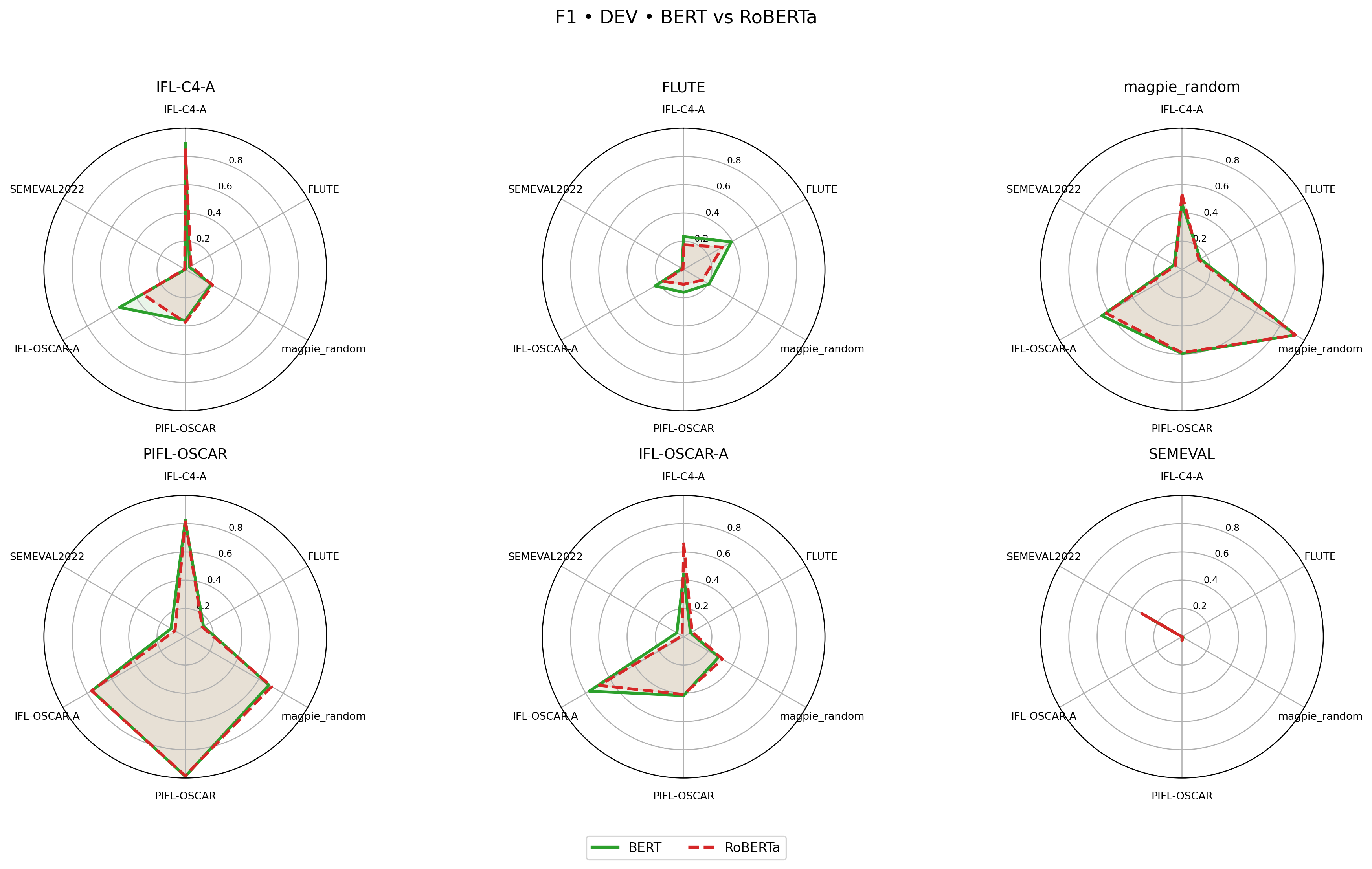}
    \caption{Cross evaluation results for F1 on test sets. BERT and RoBERTa models trained with each dataset are evaluated on the dev set of all six datasets.}
    \label{fig:dev_f1}
\end{figure}

\begin{figure}
    \centering
    \includegraphics[width=1\linewidth]{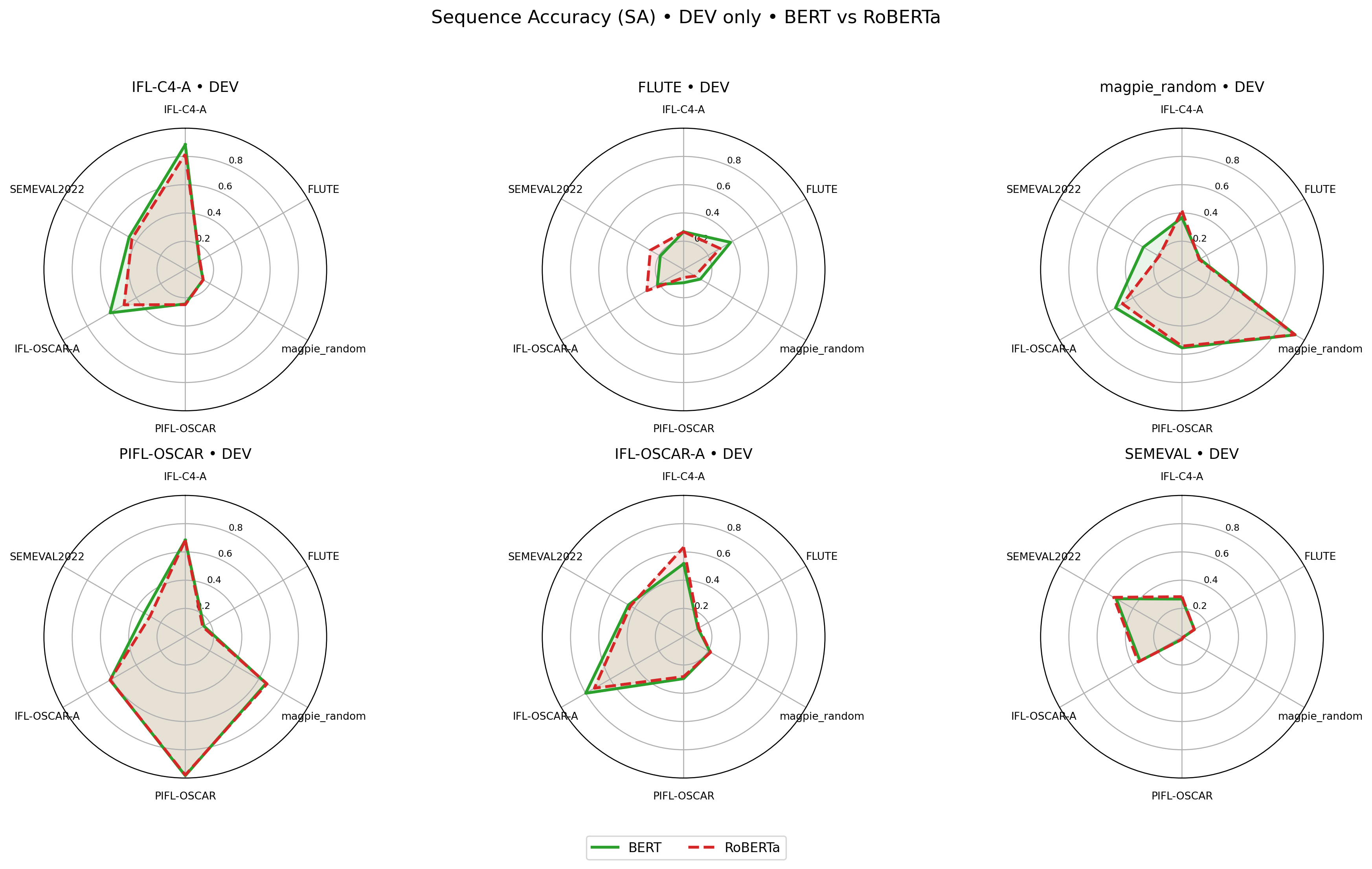}
    \caption{Cross evaluation results for sequence accuracy on test sets. BERT and RoBERTa models trained with each dataset are evaluated on the dev set of all six datasets.}
    \label{fig:dev_sa}
\end{figure}
\bibliography{ref}
\end{document}